\documentclass[acmsmall]{acmart}

\setcopyright{none}
\renewcommand\footnotetextcopyrightpermission[1]{}

\usepackage{array}
\usepackage{wrapfig}
\usepackage[pagewise]{lineno}
\usepackage{caption}
\usepackage{subcaption}
\usepackage{graphicx}
\usepackage{xcolor}

\AtBeginDocument{%
  \providecommand\BibTeX{{%
    \normalfont B\kern-0.5em{\scshape i\kern-0.25em b}\kern-0.8em\TeX}}}

\newcommand{\Rajan}[1]{\textbf{\sffamily{\textcolor{teal}{[#1 -- Rajan]}}}}
\newcommand{\DD}[1]{\textbf{\sffamily{\textcolor{red}{[#1 -- Devleena]}}}}
\newcommand{\SC}[1]{{\sffamily{\textcolor{purple}{[#1 -- Sonia]}}}}
\newcommand{\thomas}[1]{{\sffamily{\textcolor{blue}{[#1 -- Thomas]}}}}

\acmJournal{TIIS}
\acmVolume{00}
\acmNumber{00}
\acmArticle{00}
\acmMonth{00}

\begin{document}

\title{Explainable Activity Recognition for Smart Home Systems}


\author{Devleena Das}
\affiliation{%
 \institution{Georgia Institute of Technology}
 \country{USA}}

\author{Yasutaka Nishimura}
\affiliation{%
 \institution{KDDI Research}
 \country{Japan}}
 
\author{Rajan P. Vivek}
\affiliation{%
 \institution{Georgia Institute of Technology}
 \country{USA}}

\author{Naoto Takeda}
\affiliation{%
 \institution{KDDI Research}
 \country{Japan}}
 
\author{Sean T. Fish}
\affiliation{%
 \institution{Georgia Institute of Technology}
 \country{USA}}
 
\author{Thomas Pl{\"o}tz}
\affiliation{%
  \institution{Georgia Institute of Technology}
  \country{USA}
}

\author{Sonia Chernova}
\affiliation{%
 \institution{Georgia Institute of Technology}
 \country{USA}}

\renewcommand{\shortauthors}{Das et al.}

\begin{abstract}
Smart home environments are designed to provide services that help improve the quality of life for the occupant via a variety of sensors and actuators installed throughout the space. Many automated actions taken by a smart home are governed by the output of an underlying activity recognition system.  However, activity recognition systems may not be perfectly accurate and therefore inconsistencies in smart home operations can lead users reliant on smart home predictions to wonder ``why did the smart home do that?”  In this work, we build on insights from Explainable Artificial Intelligence (XAI) techniques and introduce an explainable activity recognition framework in which we leverage leading XAI methods (LIME, SHAP, Anchors) to generate natural language explanations that explain what about an activity led to the given classification. We evaluate our framework in the context of a commonly targeted smart home scenario: autonomous remote caregiver monitoring for individuals who are living alone or need assistance. Within the context of remote caregiver monitoring, we perform a two-step evaluation: (a) utilize ML experts to assess the sensibility of explanations, and (b) recruit non-experts in two user remote caregiver monitoring scenarios, synchronous and asynchronous, to assess the effectiveness of explanations generated via our framework.  Our results show that the XAI approach, SHAP, has a 92\% success rate in generating sensible explanations. Moreover, in 83\% of sampled scenarios users preferred natural language explanations over a simple activity label, underscoring the need for explainable activity recognition systems. Finally, we show that explanations generated by some XAI methods can lead users to lose confidence in the accuracy of the underlying activity recognition model, while others lead users to gain confidence. Taking all studied factors into consideration, we make a recommendation regarding which existing XAI method leads to the best performance in the domain of smart home automation, and discuss a range of topics for future work to further improve explainable activity recognition.

\end{abstract}

\ccsdesc[500]{Human-centered computing~Empirical studies in ubiquitous and mobile computing}
\ccsdesc[500]{Human-centered computing~Empirical studies in HCI}
\ccsdesc[500]{Computing methodologies~Machine learning approaches}

\keywords{Explainable AI, Smart Home Activity Recognition, Human-AI-Interaction}

\maketitle

\section{Introduction} 

Smart homes are residential environments that are augmented with sensors, actuators and computational reasoning systems designed to provide services that improve quality of life for the occupant \cite{alam2017review}.  Research in, and deployment of, smart home environments has expanded rapidly in recent years due to the availability of low cost and low power sensors, complemented with advances in wireless technologies and generalizable Machine Learning systems \cite{gayathri2017probabilistic,fahad2021activity,liciotti2020sequential}.  While some smart home systems simply allow users to remotely control devices (e.g., change the thermostat while away from the house) \cite{ford2017categories}, of greater long-term interest and impact are smart homes that provide intelligent automated assistance.  Example capabilities range from automating the lights to complement the user’s activity \cite{makonin2012smarter}, to turning off appliances if the user leaves the house \cite{kumar2016iot}, monitoring activities of daily living \cite{fahad2021activity,ni2016context,nef2015evaluation}, alerting caregivers of anomalies in a user’s behavior \cite{bakar2016activity,ghayvat2018smart}, and providing aid to users who require assistance in independent living \cite{morris2013smart,pal2017smart}.  Through the development of such capabilities, smart homes have the potential for significant societal impact in supporting healthcare and independent living.

In each of the above examples, automated actions taken by the smart home are governed by the output of an underlying Human Activity Recognition (HAR) system. 
For example, detecting that the user is waking up may trigger the lights to turn on, leaving the house may cause the stove to turn off, and repeated visits to the medicine cabinet may result in a verbal reminder that medications were already taken that day.  However, activity recognition systems are not perfectly accurate.  
For example, in \cite{al2017activity}, 13\% of Work activities were misclassified as Relaxing, in \cite{thapa2020deep} 6\% of Brushing Teeth activities were misclassified as Sleeping, and in \cite{gochoo2018unobtrusive} 2\% of Sleeping activities were misclassified as Housekeeping.  Although performance of activity recognition algorithms will continue to improve, expectations of perfectly accurate systems are impractical, particularly in complex real-world settings in which occupant behavior or the number of occupants may change over time. 

Inconsistencies in activity recognition lead to smart home operations that can be inappropriate to the household occupant, such as turning off the lights even though the occupant is still awake. In other instances, inconsistencies in activity recognition may be surprising to remote caregivers, such as the smart home predicting that the household occupant is ``leaving home" at 2 am. To date, no research has considered the \textit{explainability} of smart home operations, and activity recognition in particular.  
Do users of smart homes find it sufficient to just view the label of the recognized activity?  
Or do they prefer a more detailed explanation of why a particular activity was detected?  If the latter, how can such explanations be generated, and what information should they contain in order to be interpretable by users who are not experts in Artificial Intelligence (AI)?

Prior work in human-computer-interaction (HCI) has demonstrated that the explainability of AI systems is important for their success \cite{xu2019toward, ehsan2020human,abdul2018trends}.  The field of Explainable AI (XAI) has emerged specifically for research on the development of interpretable machine learning algorithms that can increase the transparency of black-box models  \cite{adadi2018peeking,das2020opportunities,dovsilovic2018explainable,zhang2018visual,ribeiro2016should}. While the majority of  XAI techniques focus on expert systems designed for machine learning developers \cite{selvaraju2017grad, samek2019explainable, wu2017beyond,zhang2019interpreting}, a growing body of work at the intersection of XAI and HCI has developed explainable methodologies targeting non-technical users \cite{ehsan2019automated, das2020explainable}.

In this work, our goal is to generate natural language explanations for activity recognition systems that are meaningful to non technical experts of AI. Towards this goal, we contribute a framework for \textit{explainable activity recognition} that not only generates a label for the observed activity but also an accompanying explanation about which temporal sensory observations led to the given classification by leveraging leading XAI techniques (LIME \cite{ribeiro2016should}, SHAP \cite{lundberg2017unified}, and Anchors \cite{ribeiro2018anchors}). Specifically, we first evaluate the applicability of leading XAI techniques to activity recognition data through an expert analysis. We then examine the effectiveness of each XAI technique in the context of activity monitoring for smart home systems, focusing on a healthcare scenario in which a remote caregiver seeks to monitor the activities of an older adult living alone. Through two different studies with everyday users, we compare user understanding of smart home activity labels with respect to explanations from each XAI technique. Our work contributes several key findings:
\vspace*{-0.2em}
\begin{enumerate}
\item Everyday users significantly prefer the natural language explanations generated from our explainable activity recognition framework, in comparison to the current gold standard activity label outputs of activity recognition models.

\item In both real-time and asynchronous remote caregiver monitoring settings, our framework's explanations provide increased user confidence in correct activity recognition model predictions, specifically with SHAP-based explanations providing the most confidence.

\item In an expert analysis, the SHAP XAI-model generates the most sensible explanations for multivariate activity recognition data, in comparison to the other XAI models, but at an increased computational cost.

\end{enumerate}
\vspace*{-0.25em}

To the best of our knowledge, this is the first work that adopts leading XAI techniques (LIME \cite{ribeiro2016should}, SHAP \cite{lundberg2017unified}, and Anchors \cite{ribeiro2018anchors}) for the purposes of explaining activity recognition systems to non-AI experts. With our findings, this work is a first step towards providing explainable activity recognition for situated settings.

\section{Related Work}
\label{sec:related_work}
Our work aims at exploring the value automated explanations can bring to smart home applications and practical deployments.
In what follows we summarize existing work related to two major research directions: 
\textit{i)} activity recognition in smart homes; and 
\textit{ii)} methods of explainable AI.
Our work draws upon methods from both categories, thereby contributing to accessible and understandable, and thus potentially more acceptable, smart home technologies.


\vspace{-.3cm}
\subsection{Activity Recognition in Smart Homes}
Smart home environments are designed to provide services that help improve the quality of life for the occupant via a variety of sensors and actuators installed throughout the space, and through automated inference about an occupant's activities and their relevant contexts \cite{ricquebourg2006smart}. 
A wide variety of sensors have been utilized to capture relevant contextual factors in smart homes, including (but not limited to) those that measure ambient temperature, record sound, motion, wifi signal characteristics, visual information (through, for example, cameras), and proximity information \cite{bakar2016activity,williams2019survey,abowd2002aware}.
Furthermore, body worn sensors are also widely used for direct recording of an occupant's movements \cite{wang2011recognizing}.  
In the context of the smart home domain, activity recognition specifically serves to identify and log the occupant's activities of daily living (ADL) from temporal-ordered sensor data.  
The automatically inferred activity labels then determine which assistive actions the smart home shall perform, such as turning on a light, closing the garage door, or adjusting a thermostat.  
Activity recognition in smart home environments remains a challenging problem specifically demanding innovation in Machine Learning (ML) for automated sensor data analysis, mainly due to the diversity of environments, users, sensor configurations and actuators that must be considered, particularly due to the unique nature of each individual household and user.



Prior work in HAR for smart home domains has studied a wide variety of Machine Learning classification techniques, including Conditional Random Fields \cite{nazerfard2010conditional}, Naive Bayes \cite{ravi2005activity}, Hidden Markov Models \cite{patterson2005fine, van2008accurate}, and Artificial Neural Networks \cite{bourobou2015user, mehr2016resident}. Recently, with the popularity of deep learning, Liciotti et al. \cite{liciotti2020sequential} have shown that LSTM based architectures for smart home systems can outperform the above mentioned traditional ML techniques. Specifically Liciotti et al. employed five LSTM architectures 
(uni-LSTM, Bi-LSTM, Cascading-LSTM, Ensemble-LSTM, and Cascade-Ensemble-LSTM) and compared their performances between each LSTM model as well as against traditional ML techniques.  The authors reason that the network’s ability to model long term dependencies between sequences of data and to learn non linear feature representations allow for the LSTM-based models to better model ADL patterns in unbalanced datasets. 

Several large-scale smart home activity recognition datasets have been developed to aid the research community in robust evaluation and benchmaking. Datasets capturing activities of daily living include CASAS \cite{cook2012casas}, ARAS \cite{alemdar2013aras}, Placelab \cite{intille2005placelab}, the MIT Activity Recognition dataset \cite{tapia2004activity}, and the van Kasteren dataset \cite{van2008accurate}. Each of these datasets covers a varying range of ADL activities, and differs in the number of household occupants, as well as the number and types of sensors utilized for data collection. For example, the ARAS dataset includes multiple household occupants observed over a two month time-span using 20 home sensors, while the CASAS dataset covers a range of both single and multi-person households over 2-8 month time spans using 20-86 sensors.

In this work, our goal is not to improve the state-of-the-art in activity recognition itself.  Instead, our work explores the added benefit that model-agnostic XAI methods can bring when paired with leading HAR approaches.  
To do so, we leverage Liciotti et al.'s state-of-the-art LSTM-based HAR \cite{liciotti2020sequential}, specifically adapting their uni-LSTM model.  We validate our approach using the CASAS dataset, specifically using the Milan household data, which was also used in \cite{liciotti2020sequential}. We describe our model and the dataset in detail in Section \ref{sec:AR-LSTM}.

\subsection{Explainable AI (XAI)}

The field of Explainable AI focuses on the development of algorithms that provide insights into how an AI system makes decisions and predictions.  A number of recent surveys of XAI provide detailed perspectives on various aspects of the XAI problem, including general surveys of the XAI field \cite{adadi2018peeking,das2020opportunities,dovsilovic2018explainable,zhang2018visual,ribeiro2016should}, XAI applied to medical domains \cite{tjoa2020survey}, use of natural language processing in XAI \cite{qian2021xnlp}, and user experience in XAI research practices \cite{ferreira2020people}.  Due to the breadth of the field, we focus our discussion here on XAI methods designed for classification-based problems, such as activity recognition as it is targeted by our work, as well as XAI methods addressing non-expert users.  We refer readers to the above surveys for all additional topics relating to XAI.



\noindent
\paragraph{Insights from psychology:}
Prior work in the field of psychology has provided insights into the desirable qualities of an explanation. Early work by Hilton \cite{hilton1990conversational} posed explanations as a social and conversational process, arguing that good explanations must not only be true, but must answer a ``why'' question (whether one was asked or not).  A taxonomy of explanations used in a number of psychology works \cite{dennett1989intentional,lombrozo2012explanation} further categorizes explanations into three types:  
\textit{i)} mechanistic; 
\textit{ii)} teleological; or 
\textit{iii)} formal. 
Mechanistic explanations are given with regards to how something functions, and teleological explanations appeal to purpose. These types of explanations are helpful for explaining processes. Formal explanations are those which are concerned with categorical definition and help explain why an element is considered as part of a set. As our work seeks to improve non-expert understanding of a classifier's decision, our explanations are most concerned with mechanistic explanations. 



\noindent
\paragraph{Explaining classification problems: }
Classification-based XAI methodologies can be categorized along several axes. First we categorize XAI techniques by the complexity and generalizability of models they can explain:
\begin{description}
    \item [model-intrinsic,] which refer to models that leverage an inherently interpretable structure, such as a rule list \cite{letham2015interpretable} or decision tree \cite{tan2020tree}, that does not require further processing to be explainable; or
    
    \item [model-agnostic,] such that they provide explanations for underlying computational models that in themselves are not interpretable \cite{zhao2021causal, ribeiro2016should,lundberg2017unified,ribeiro2018anchors}.
    
    
\end{description}
Model-intrinsic techniques do not scale well to complex, multi-dimensional spaces \cite{adadi2018peeking}. Model-agnostic techniques are typically more desirable since they are not dependent upon a fixed underlying classification model and can be applied widely to any state-of-the-art technique.


    

Additionally, XAI techniques can be characterized as:
\begin{description}
    \item [local,] which explain the behavior of the model for a specific singular decision; or 
    
    \item [global,] which explain the behavior and reasoning of the entire model.
\end{description}
In the context of activity recognition we are most interested in local models that explain the model reasoning that led to the generation of a specific activity label.  
For example, if a smart home turns off the lights while the user is watching TV, the user may ask ``why did you think I was sleeping?" and will require an answer about the classification of that particular activity and not a description of the complete HAR model.

In this work, we focus on XAI techniques that are designed to be model-agnostic and provide local interpretability.  
State-of-the-art model-agnostic methodologies include LIME \cite{ribeiro2016should}, SHAP \cite{lundberg2017unified}, and Anchors \cite{ribeiro2018anchors}. While LIME utilizes perturbation to find surrogate models that fit and explain individual instances, SHAP leverages game theory foundations of Shapley Values \cite{shapley201617}, which provide the marginal contributions of features towards an input instance being explained. Anchors formulates a multi-armed bandit problem \cite{kaufmann2013information} found in reinforcement learning to produce ``IF-THEN” rules that provide local interpretability for individual instances.
Each of these methodologies outputs an ordered list of the  local features most relevant to a classification (e.g., "[M026, M017, T002]").  In this paper, we examine the effectiveness of LIME, SHAP and Anchors when applied to activity recognition for smart home data, and contribute techniques for generating interpretable natural language explanations from their output.  We provide additional details of each of these algorithms and our approach in Section \ref{sec:xai-models}.


\smallskip
\textit{XAI for everyday people: }
Most prior work in XAI has focused on developing explantion techniques for machine learning experts, developers, or specific domain experts \cite{adadi2018peeking, samek2019explainable, gade2019explainable, selvaraju2017grad, zhang2019interpreting}. However, recent efforts at the intersection of XAI and HCI have led to increased development of techniques for interaction with everyday users, or non-experts. 
For example, in the context of explaining video activity recognition for cooking tasks, Nourani et al. \cite{nourani2020don} establish that high levels of veracity in explanations can improve human task performance and system understanding. Additionally, recent work has established the importance of contextualized reasoning as well as the importance of natural language explanations for effective understanding. Specifically, Ehsan et al.\  \cite{ehsan2019automated} leveraged sequence-to-sequence learning to autonomously generate natural language rationales in the domain of Markovian-based games, such as Frogger, to study user preferences of rationales. Their results demonstrate that users significantly prefer ``complete-view" rationales, which utilize the entire state space as context, as opposed to ``focused-view", which utilize only a subset of the full state space. 
Additionally, in the specific context of understanding robot failures yet related to the work presented here,  Das et al.\  \cite{das2020explainable} extended sequence to sequence learning to autonomously generate context-based natural language explanations in a continuous state space. 
Their results demonstrated that explanations grounded in environmental as well as temporal and / or historical context, most accurately helped non-experts identify robot failures as well as correct recovery solutions. However, both of the above sequence-to-sequence methodologies require vast amounts of training data in the form of expert-labeled explanations, making them challenging to generalize to new applications.  
In this work, we take a template-based approach to generating explanations, similar to Elizalde et al.\  \cite{elizalde2009generating} who utilize template-based approach to explain recommendations made by a Markov Decision Process for operating a steam generator. 

\smallskip
\textit{Evaluation of explanations:} A recent survey by Miller \cite{miller2019explanation} provides valuable insights on how computational techniques from explainable AI can build on existing research in social sciences, reviewing relevant papers from philosophy, cognitive psychology/science, and
social psychology, and how they relate to XAI.  Of particular interest is Miller's investigation of the criteria that researchers have used to evaluate explanations, which include the coherence and simplicity of an explanation \cite{thagard1989explanatory,read1993explanatory}, as well as its truthfulness \cite{kulesza2013too}.  In this work, we evaluate our explanation for truthfulness in Section \ref{sec:xai_exp_analysis}.  We then control for truthfulness, and evaluate the coherence of our explanations with real-world users in Section \ref{sec:user-eval}.

\section{Overview}
\label{sec:overview}

Our overarching goal is to introduce an explainable activity recognition framework 
in which explainable AI techniques can be integrated with activity recognition to make activity recognition models interpretable to everyday people through natural language explanations. We explore the benefit of our explainable activity recognition framework in the context of a commonly targeted smart home scenario: autonomous remote monitoring for individuals who live alone or need assistance. While most of prior work focuses on methods for robust sensor data collection for remote monitoring \cite{shahjalal2020overview, zhai2011design, rialle2002health}, we focus on the scenario of providing remote caregivers with meaningful explanations of a household occupant's activities as a means of monitoring the occupant's wellbeing.

To evaluate our framework in the context of remote caregiver monitoring, we utilize a widely used smart home activity recognition dataset (the CASAS Milan dataset \cite{cook2012casas}) and automatically generate natural language explanations of activity labels using our explainable activity recognition framework.  Specifically,  we take a leading activity recognition model (the uni-LSTM from \cite{liciotti2020sequential}) and integrate it with three leading model-agnostic XAI techniques -- LIME \cite{ribeiro2016should}, SHAP \cite{lundberg2017unified} and Anchors \cite{ribeiro2018anchors}.\footnote{We additionally introduce LIME+ in Section \ref{sec:lime+}, as an extension of LIME \cite{ribeiro2016should}, to investigate the benefits of duration information in explanations.}

Across these efforts, our work seeks to answer the following research questions:
\begin{description}
    \item [RQ1:] Do smart home users prefer natural language XAI-model based explanations over simple activity labels?
    
    \item [RQ2:] Do XAI-model based explanations give users more confidence in the activity recognition model?
    
    \item [RQ3:] Out of the leading XAI methods explored in this work, which one produces the most sensible explanations for smart home activity recognition models?

\end{description}

Within the context of remote caregiver monitoring, we validate our research questions through a two-step 
evaluation. First, we evaluate for explanation sensibility via an expert analysis. Second, we perform two IRB-approved user studies regarding remote smart home monitoring for caregivers, and evaluate the effectiveness of each explanation in helping non-technical users understand activity labels from a smart home system. In the subsections below, we present an overview of our approach, including the core components of our framework and our evaluation method.

\textbf{To the best of our knowledge, this is the first work to consider explainability of activity recognition models for non-expert users, and answering the questions above in relation to \textit{established} XAI techniques will serve to guide the development of \textit{novel} XAI techniques in this area in future work.}


\begin{figure}
\centering
\includegraphics[width=1.0\columnwidth]{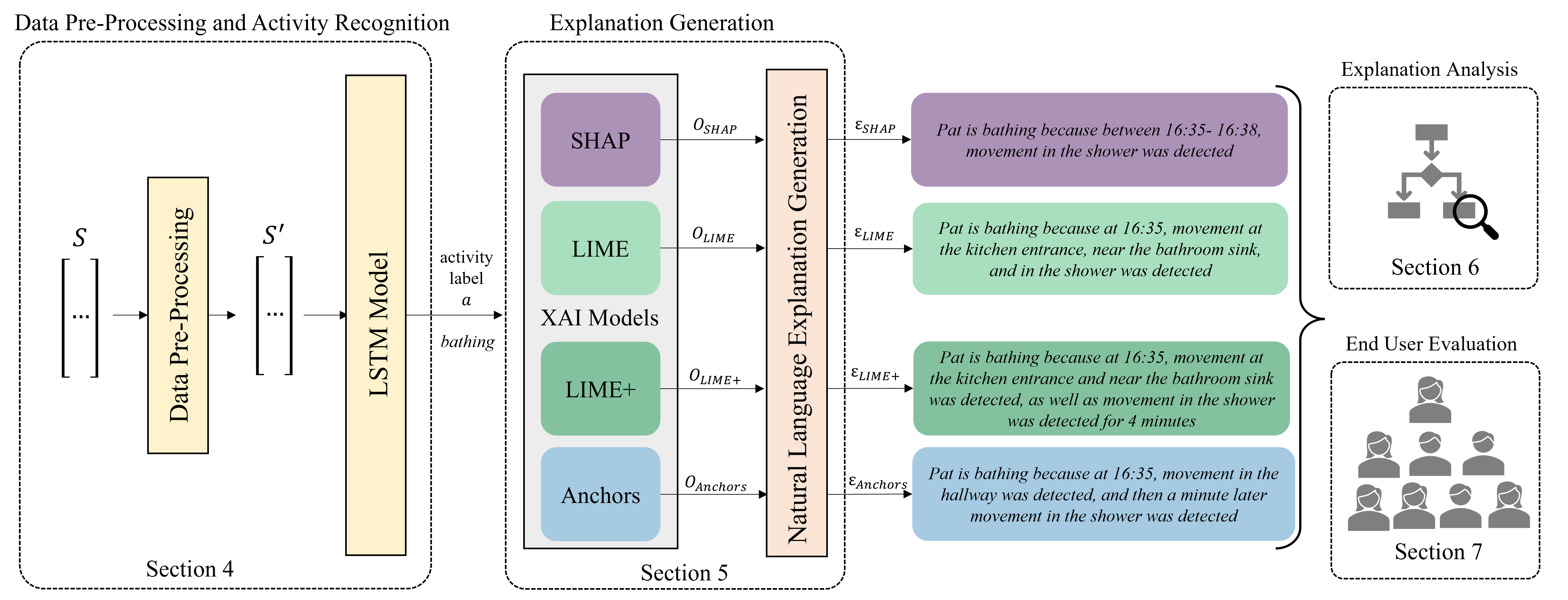}
\caption{Overview of the work presented in this paper, outlining the different components within our explainable activity recognition framework. For data pre-processing and activity recognition (Section \ref{sec:AR-LSTM}), we pre-process smart home data $S$ into a multivariate, fixed-interval representation $S^{'}$, suitable for explainable activity recognition. The output activity label, $a$, as well as the activity recognition model is used by each XAI model to generate top contributing features $O$. These features are translated into a natural language explanation $\mathcal{E}$, corresponding to the XAI model (Section \ref{sec:xai-models}). We analyze the sensibility of the explanation types via an expert analysis (Section 6) as well as their effectiveness in helping end users understand a smart home's activity recognition label (Section \ref{sec:user-eval}).}
\label{fig:overview_diag}
\end{figure}

\subsection{Problem Formulation}
\label{sec:overview:problem}
Let $S=\{(s_1,v_1,k_1),...,(s_{t},v_{t},k_{t})\}$ represent a sequence of sensor events of length $t$, where sensor $s_i$ takes on the value $v_i$ at time $k_i$.  Let $A=\{a_1,...,a_{n}\}$ represent the set of activity labels.  Given this data, our goal is to find model $f(S) \rightarrow \{a,\mathcal{E}\}$, where $a\in A$ represents the activity label for the events described by $S$, and $\mathcal{E}$ is an explanation that describes which events in $S$ most significantly influenced the selection of $a$ as the label.  As in prior work on XAI for non-technical users \cite{ehsan2019automated, das2020explainable}, we seek to represent $\mathcal{E}$ in the form of a natural language phrase.  

\subsection{Explainable Activity Recognition Framework}
Figure \ref{fig:overview_diag} gives an overview of our approach for the above problem. The objective of the first component of our framework, described in Section \ref{sec:AR-LSTM}, is to pre-process $S$ in a manner that represents the information needed to generate smart home explanations preferable to everyday users. We first conducted a user study that evaluated user preferences towards the types of explanations producible from $S$. Given the study results, we then pre-process $S$ into a fixed-interval multivariate data representation, $S'$, that facilitates explainable activity recognition. We then use $S'$ to generate an activity label $a$ using an LSTM model extended from \cite{liciotti2020sequential}.

The second component of our framework, presented in Section \ref{sec:xai-models}, generates the natural language explanations, $\mathcal{E}$, that describe the features of $S'$ that most significantly contributed to the classifier's decision. Our work evaluates four different techniques for explanation generation based on state-of-the-art techniques in XAI. Specifically, we extend the following three leading XAI methods from the literature:

\begin{description}
    \item  [Local Interpretable Model-agnostic Explanations (LIME) \cite{ribeiro2016should}] -- a model-agnostic XAI technique that explains the predictions of a black-box classifier by learning an interpretable model locally around the prediction;
    \item [Anchors \cite{ribeiro2018anchors}] -- a model-agnostic XAI technique that explains the behavior of black-box models with high-precision rules found via a multi-arm-bandit search; and
    \item [SHapley Additive exPlanations (SHAP) \cite{lundberg2017unified}] -- a  model-agnostic XAI technique that calculates Shapley values \cite{shapley201617} to understand the feature importance of all features utilized by a black-box model for a prediction.
\end{description}

Each method is used to generate an explanation $\mathcal{E}_{LIME}, \mathcal{E}_{Anchors}$ and  $\mathcal{E}_{SHAP}$, respectively.  
Additionally, we extend the LIME methodology to introduce \textbf{LIME+} and $\mathcal{E}_{LIME+}$ explanations, which improve upon LIME by identifying blocks of time (rather than individual timesteps) during which sensors contribute strongly to an instance's classification, allowing for more temporal and intuitive explanations. We selected these techniques based on their XAI property of being model agnostic and applicable to any black-box model used for smart home activity recognition, as well as their ability to provide intuitive outputs that can be translated into natural language explanations for end users. Given the resulting four explanation types, we compare their performance along multiple relevant metrics.

\subsection{Evaluating XAI-Model Based Explanations}
In order to understand the effectiveness of each of the above explanation types, we first perform an expert evaluation of the sensibility of generated XAI-model based explanations.
To achieve this, first we develop a classification rule set that determines whether an explanation is sensical or nonsensical based on the sensor layout of the smart home in which activities were performed. In Section \ref{sec:xai_exp_analysis} we compare the sensibility of explanations across each explanation type, as well as examine the computational costs and feature distributions of each XAI model.

Second, we analyze the effectiveness of each explanation type in helping end users understand a smart home's activity recognition label in monitoring the well being of older individuals living alone. Specifically, utilizing only the categorized sensible explanations\footnote{Note that we use only the sensible explanations in our user study to evaluate user perceptions of each explanation type without the confounding variable of explanation sensibility.}, we conduct two IRB-approved user studies to evaluate user preferences toward each explanation type and their perceptions of the smart home system's capabilities. 
Note, while remote caregiver monitoring may arise in many different forms, we categorize remote monitoring into real-time or asynchronous scenarios. For example,
a caregiver may be hired to monitor, in real-time, the well-being of multiple higher-risk individuals. Alternatively, remote caregiver monitoring may involve an adult child reviewing an elderly parent's daily activities in a lower-risk setting. Our user studies emulate these two scenarios.
In the first user study, we perform a Wizard of Oz evaluation in which participants are asked to remotely monitor the real-time activities of four simulated, higher-risk individuals living alone. In the second user study, we recruit participants at a large scale from Amazon Mechanical Turk (AMT) to role-play as a remote caregiver and asynchronously monitor the activities of another simulated individual. We present these studies in Section \ref{sec:user-eval}.


\section{Activity Recognition for Explainable Smart Homes}
\label{sec:AR-LSTM}
Activity recognition represents the basis for the work presented in this paper.
We adopt a state-of-the-art approach \cite{liciotti2020sequential} in the smart home domain, and adapt it towards our application scenario of generating meaningful smart home explanations for everyday users. In this section, we first present the user study we conducted to evaluate user preferences towards plausible smart home explanation types, given a sequence of sensor events  $S$.

Informed by the results of the study, which show that users exhibit preference for temporal and sensor-related information,
we then introduce the multivariate, fixed interval data representation we use for activity recognition. We then describe our adapted activity recognition model and evaluate its overall effectiveness in recognizing relevant activities from analyzing sensor data as they are captured in a typical smart home.

\subsection{User Preference of Explanation Types}
\label{sec:user-pref-study}
Since our work is one of the first efforts towards generating explanations for non-expert users within the smart home community,
one of the first questions to answer is, "What types of explanations do non-expert users find helpful in understanding smart home behavior?" Prior work demonstrates that taking a design-centric approach, in which we first understand non-experts' preferences towards explanations in a given domain, can aid in generating more meaningful explanations for our target audience \cite{das2020explainable,ehsan2021explainable}.  Thus, we begin by presenting the results of a user study in which we compare two explanation styles for activity recognition data.

Our explanations extend the representation used in the state-of-the-art smart home activity recognition model by Liciotti et al. \cite{liciotti2020sequential}, in which importance of smart home activities is communicated through the sequential ordering of sensor events (e.g., fridge sensor activated, then fridge sensor closed, then sink sensor activated). This sequential ordering enables the user to follow the sequence of events that occurred in an \textit{ordered} fashion. However, this representation does not communicate the \textit{temporal duration} of each event.  Prior work within the field of activity recognition has shown the importance of incorporating temporal duration information to improve activity recognition performance \cite{ma2019attnsense, zeng2018understanding,yao2018qualitydeepsense}, and we hypothesize that humans would similarly benefit from temporal information. Thus, to understand what information users prefer to have we conducted a user study comparison between the following two explanation types:

\begin{itemize}
    \item \textit{Ordered Non-Temporal:} sequential ordering of 
    the top most important
    sensor events determined by an XAI algorithm (see Section \ref{sec:xai-models}). An example is, ``The activity is `bathing' because movement near the bathroom sink was detected, movement near the shower was detected, and movement near the shower was detected." 

    \item \textit{Ordered Temporal:} sequential ordering of the top most important sensor events, determined by an XAI algorithm (see Section \ref{sec:xai-models}),
    supplemented with temporal data in the form of both absolute time and duration. An example is, ``The activity is `bathing' because between 06:39 - 06:44 movement near the shower was detected."
\end{itemize}
Note, given the absence of temporal duration information in \textit{Ordered Non-Temporal} explanations, sequential sensor events from the same sensor are not grouped as they are in \textit{Ordered Temporal} explanations. In Table \ref{temp-nontemp-exps} we provide additional explanations of \textit{Ordered Non-Temporal} and \textit{Ordered Temporal} explanations for different smart home activities and highlight the differences between these explanation types discussed above. The explanations are generated via XAI-models later described in Section \ref{sec:xai-models}. Note that the features utilized in the \textit{Ordered Non-Temporal} and \textit{Ordered Temporal} explanations differ. This effect is due to the XAI-models calculating feature importance across the differing data representations. Recall, \textit{Ordered Temporal} includes both temporal data as well as sensor events, whereas \textit{Ordered Non-Temporal} only includes sensor events.

For our user study, we recruited 50 participants from Amazon Mechanical Turk, all of whom had an approval rate greater than 99\%. Specifically, our participants included 29 male and 19 female, all of whom were 18 years or older (M=38.1, SD=10.1). The study took, on average, 15 minutes and participants were compensated US \$2.50. To reduce participant fatigue, 25 randomly selected participants were presented with 24 scenarios of activities performed by a simulated individual Pat, while the remaining 25 participants were presented with another 24, different scenarios. Each set of 24 scenarios was randomly sampled, without replacement, from a pool of 93 scenarios. For each scenario, participants were asked to play the role of a remote caregiver, monitoring the well-being of an elderly patient with Dementia named Pat. Participants were presented with Pat's activity (e.g., “Pat is currently bathing.").  Participants were then shown two explanations and asked to select which explanation they would find most helpful in understanding the smart home's activity recognition label for Pat's activity. If both explanations were equally preferable, users could select "No preference".

\begin{figure}[t]
\centering
\includegraphics[width=0.43\columnwidth]{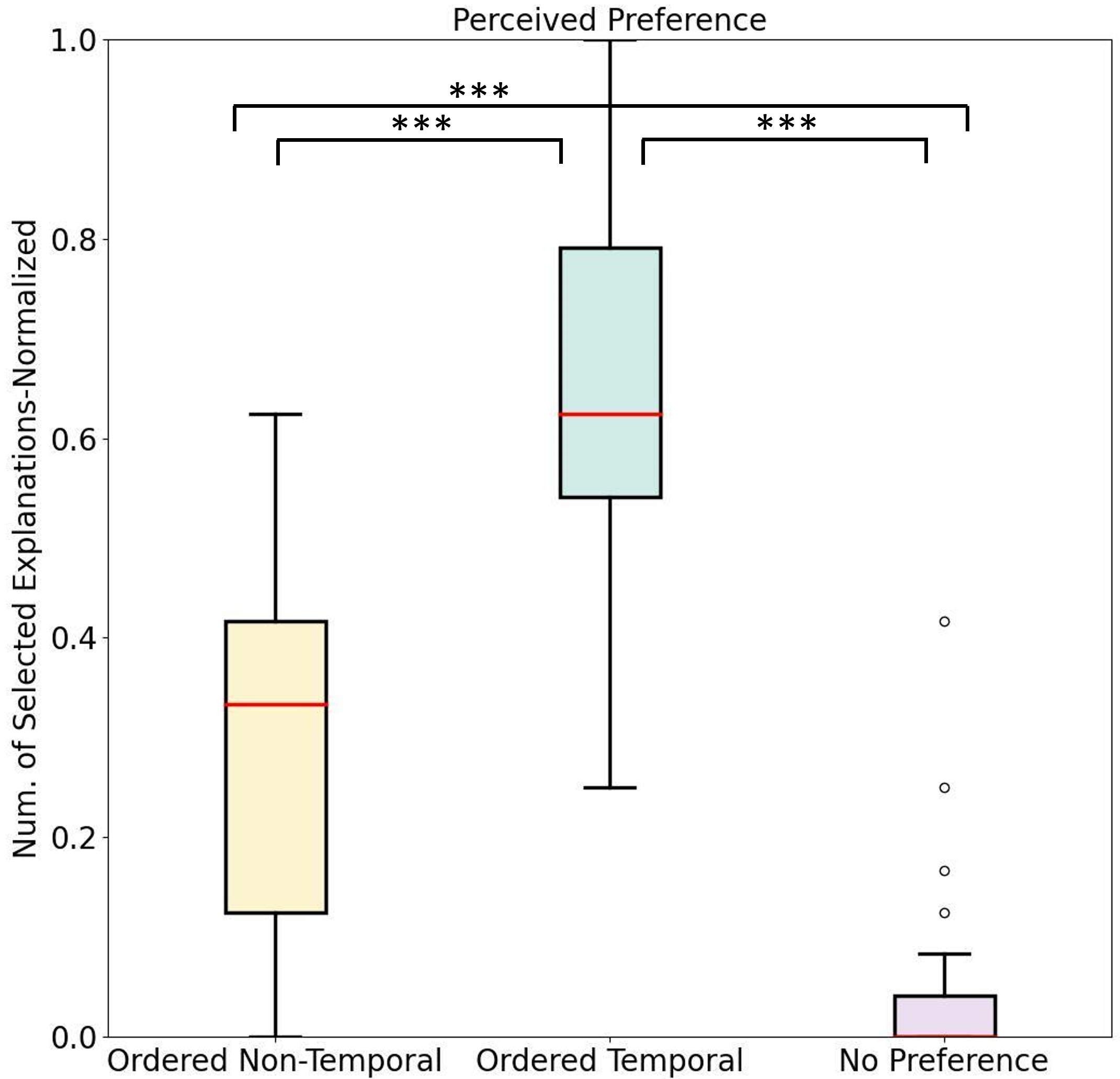}
\caption{Comparison of user selected preferences towards explanation types that differ by the exclusion or inclusion of temporal information. Statistical significance reported as *** p<0.001.}
\label{fig:exp_pref_study}
\end{figure}

In Figure \ref{fig:exp_pref_study}, we present the participants' selected preferences across all 48 scenarios\footnote{We performed additional analyses on the two groups individually to ensure no grouping bias. Please see Appendix \ref{sec:appendix-study1}.}. Since the data did not follow a normal distribution, we utilized a non-parametric Friedman Test with a post-hoc Nemenyi test to evaluate statistical significance. We observe that overall participants did have a significant preference over the explanation styles ($\chi^2$(2) = 79.14, $p$ <0.001), and that the temporal based explanations were significantly more preferable than non-temporal based explanations ($p$<0.05). These results demonstrate the favored importance of including temporal information, in addition to sensor-related information, in a smart-home explanation and guide our data pre-processing in the following section.

\begin{table}[t]
\centering
\caption{Example examples of ordered non-temporal and ordered temporal explanations for different smart home activities.}
\label{temp-nontemp-exps}
{\small
\begin{tabular}{|>{\centering\arraybackslash}p{2.5cm}|p{4.2cm}|p{4.2cm}|}
\hline
\textbf{Activity}  &
  \multicolumn{1}{c|}{\textbf{Ordered Non-Temporal}} &
  \multicolumn{1}{c|}{\textbf{ Ordered Temporal}} \\ \hline
\textbf{Taking Medicine} &
    \small  The activity is `taking medicine' because the pantry door was closed, and then movement near the pantry was not detected and then movement near the pantry was detected. &
  \small The activity is `taking medicine' because at 08:24 the pantry door was open, movement in the kitchen was detected and movement near the pantry was detected. \\ \hline
\textbf{Leaving Home} &
\small   The activity is `leaving home' because the front door was open, and then movement near the front door was not detected and then movement near the front door was detected. &
 \small The activity is `leaving home' because at 13:13 the front door was open and between 13:13 - 13:14 movement near the front door was detected. \\ \hline
\textbf{Working} &
\small  The activity is `working' because movement near the desk was detected, and then movement near the desk was detected and then movement on the TV room couch was not detected. &
  \small The activity is `working' because between 16:35 - 16:38 movement near the desk was detected. \\ \hline
\textbf{Sleeping} &
\small The activity is `sleeping' because movement in the bedroom was detected, and then movement near the bed was detected and then movement near the bed was detected. &
 \small  The activity is `sleeping' because at 23:32 movement near the pantry was not detected, 12 minutes later movement in the TV room was not detected and then 12 minutes later movement in the bedroom was detected. \\ \hline
 \textbf{Cooking} &
\small  The activity is `cooking' because movement at the kitchen entrance was detected, and then movement at the kitchen entrance was detected and then movement in the kitchen was detected. &
 \small  The activity is `cooking' because at 13:16 movement at the kitchen entrance was detected, a minute later movement near the fridge was detected and then 2 minutes later movement near the fridge was detected.\\ \hline
 \textbf{Bed to Toilet} &
\small  The activity is `bed to toilet' because movement in the bathroom was detected, and then movement in the bathroom was detected and then movement in the bedroom was not detected. &
 \small  The activity is `bed to toilet' because at 06:37 movement near the bed and movement in the bathroom was detected and then a minute later, movement in the bedroom was detected. \\ \hline
\end{tabular}
}
\vspace*{-1em}
\end{table}

\subsection{Data Processing}
\label{sec:data-proc}
The objective of our work is to pre-process smart home data in a way that allows for generating meaningful smart home explanations for everyday users. Recall (from Section \ref{sec:overview:problem}) that $S=\{(s_1,v_1,k_1),...,(s_{t},v_{t},k_{t})\}$  represents the sensor event sequence obtained from a smart home over $t$ timesteps.  An example of such a sequence is:
\begin{quote}
\center
    (M024, 1, 03:38:28)\\
    (M024, 0, 03:45:17)\\
    (D002, 121.4, 03:50:01)
\end{quote}

\noindent
where the first value encodes the identity of the sensor, the second value corresponds to the value of the sensor, and the third value encodes the time of the event.  The value encoded by each sensor depends on the sensor type; for example, motion sensors return binary values (lines 1 and 2), while distance sensors mounted near the door may provide real-valued output (line 3). 

Given the results from Section \ref{sec:user-pref-study}, we reformat $S$ into a fixed-interval, multivariate representation that includes both temporal and sensor information.
Specifically, instead of recording sensor event changes, we use a representation that encodes the explicit value of each sensor at each timestep.  Thus, given $M$ environmental sensors and a duration of $T$ timesteps, we construct a $T \times M$ matrix $S^{'}$ such that $S^{'}[t][m]$ represents the value of sensor $m$ at timestep $t$.  Through this multivariate representation, we are able to preserve both the temporal context and sensor events for an activity. 

Additionally, characterizing smart home data in a representation that is interpretable to humans is crucial for generating human understandable explanations \cite{ribeiro2016should}. Binary sensors already possess such interpretability as they only represent a single ``on" or ``off" state. However, continuous sensor values do not inherently possess interpretability and often require a post-hoc analysis to understand its patterns and meaning. Therefore, to characterize smart home data in an interpretable manner, we discretize continuous sensor values into categorical values that represent the defining pattern observed over an interval (e.g., 1 minute).  For example, in the case of the distance sensor above, we categorize its output as ``door open''.  Similarly, if the temperature fluctuated within a one minute interval but increased at the end of the interval, then we categorize the temperature to have been ``increased".

\begin{figure}[t]
\centering
\includegraphics[width=0.9\columnwidth]{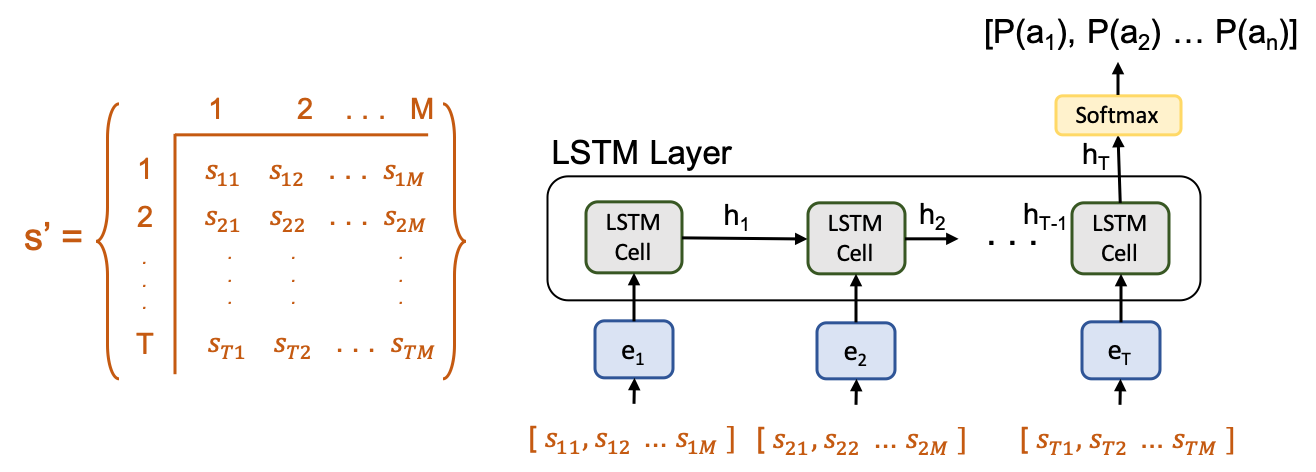}
\caption{LSTM model architecture used for activity recognition in smart homes, where $S^{'}$ represents the input data into the LSTM model, and the output is a probability distribution over all candidate activities.}
\label{fig:LSTM}
\end{figure}

\subsection{Model Architecture} 
Figure \ref{fig:LSTM} shows our adapted uni-LSTM model from  Liciotti et al. \cite{liciotti2020sequential}. The LSTM model takes in as input $S^{'}$, where each
 $S^{'}[t][1:M]$ represents a sequence of sensor events at a particular timestep $t$. Each sequence  $S^{'}[t][1:M]$ is first embedded by an embedding layer $e_{t}$ before being passed into the LSTM layer. The output of the LSTM layer is a hidden state $h_{T}$ which is passed through a dense layer with a softmax activation to obtain a probability distribution \{$P(a_{1}), P(a_{1})...P(a_{n})\}$ over all activities in $A$. The most probable activity $a \in A$ is selected as the output of the LSTM model. 


\begin{figure}
\centering
\includegraphics[width=0.4\columnwidth]{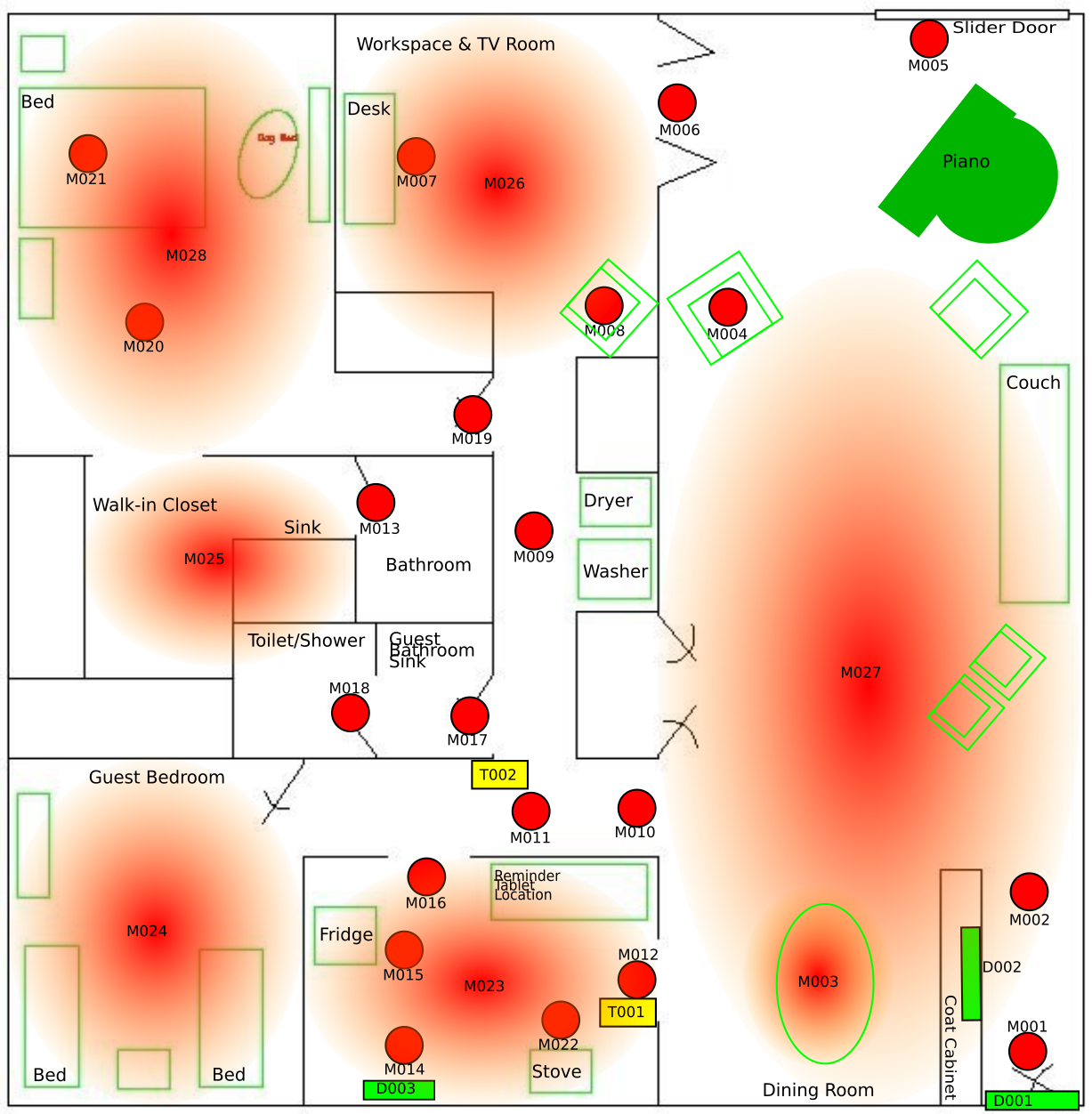}
\caption{Home Layout and Locations of Sensors utilized in the CASAS Milan Dataset \cite{cook2012casas}, where door sensors are designated as D\#\#\# (solid green boxes), motion sensors as M\#\#\# (red circles) and temperature sensors as T\#\#\# (solid yellow boxes). Each red dot represents a motion sensor that can detect motion in a localized area, whereas each radiating red area represents a motion sensor that can detect motion over the indicated area.}
\label{fig:home_layout}
\end{figure}

\subsection {Dataset}
\label{milan-dataset}

The explorations presented in this paper are generic for smart home application scenarios.
Yet, for practical considerations our developments and evaluations are based on an existing, established benchmark dataset from the field: the CASAS Milan dataset \cite{cook2012casas}. The Milan dataset contains sensor data collected from a smart home over a period of 92 days. 
Figure \ref{fig:home_layout} shows the home layout and locations of the set of 33 sensors in the home. 
The set of sensors include 3 door sensors, 28 motion sensors and 2 temperature sensors.  
The Milan dataset includes both the start and end time of an activity and includes data for 15 activities of daily living (ADL): \textit{Bed\_to\_Toilet, Chores, Desk\_Activity, Dining\_Rm\_Activity, Eve\_Meds, Guest\_Bathroom, Kitchen\_Activity, Leave\_Home, Master\_Bathroom, Meditate, Watch\_TV, Sleep, Read, Morning\_Meds, Master\_Bedroom\_Activity}. 

To remain consistent with Liciotti et al.\  \cite{letham2015interpretable}, we similarly map the 15 ADL activities onto 10 activites. Particularly, our activity set $A$ = \{\textit{Other, Work, Take medicine, Sleep, Relax, Leave home, Eat, Cook, Bed to toilet, Bathing}\}. Additionally, while the duration of an activity can vary, we set the number of timesteps representing an activity to be 30 one-minute intervals, after verifying that most activities under the Milan dataset are under 30 minutes, and that utilizing a larger duration for an activity does not affect the activity recognition performance. Thus from our data processing methodology (Section \ref{sec:data-proc}), our final dataset $D$, includes 3,298 samples of activities $a \in A$, each represented by $S^{'}$, array of size $T$ timesteps $\times$ $M$ sensors, where $T = 30$ and $M= 33$.


\begin{figure}
\centering
 \includegraphics[width=0.6\columnwidth]{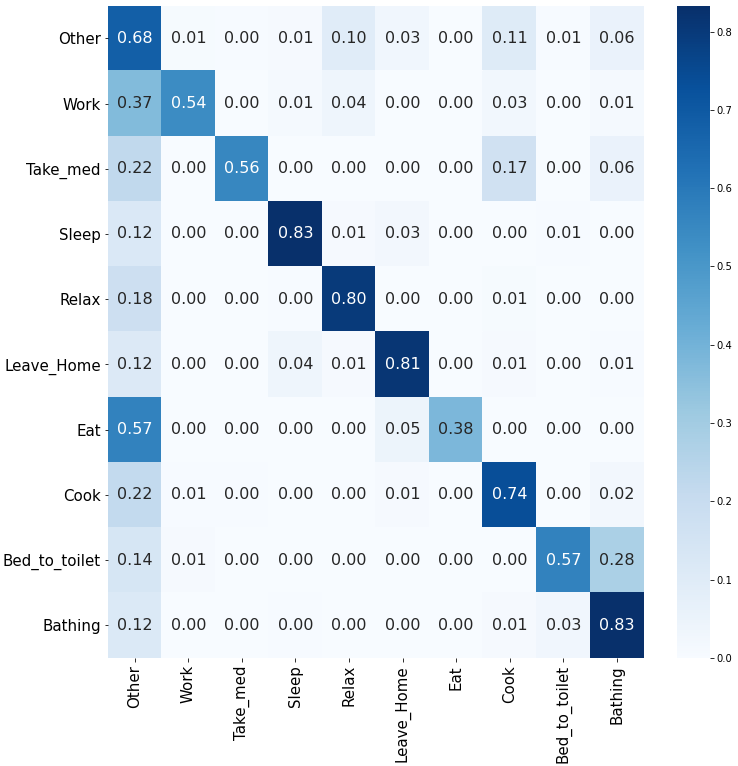}
\captionof{figure}{Confusion matrix our adapted LSTM network architecture, trained on the CASAS Milan dataset, where the y-axis denote the ground truth activities, and the x-axis denote predicted activities.}
\label{fig:confusion_matrix}
\end{figure}

 \begin{table}[t]
    \begin{minipage}{0.54\textwidth}
     \centering
    \caption{Model performance results, evaluated on CASAS Milan Dataset.}
    \label{table:modelperformance}
    \begin{tabular}{ |c|c|c|c| } 
     \hline
     \textbf{Activity} & \textbf{Precision} & \textbf{Recall} & \textbf{F1-Score} \\ 
     \hline
     Other & 0.71 & 0.68 & 0.70 \\
     \hline
     Work & 0.71 & 0.54 & 0.61 \\
     \hline
     Take medicine & 0.74 & 0.56 & 0.63 \\
     \hline
     Sleep & 0.84 & 0.83 & 0.83 \\
     \hline
     Relax & 0.72 & 0.80 & 0.76 \\
     \hline
     Leave home & 0.72 & 0.81 & 0.76 \\
     \hline
     Eat & 0.67 & 0.38 & 0.48 \\
     \hline
     Cook & 0.69 & 0.74 & 0.71 \\
     \hline
     Bed to toilet & 0.64 & 0.57 & 0.60 \\
     \hline
     Bathing & 0.81 & 0.83 & 0.82 \\
     \hline \hline
     Average & 0.73 & 0.67 & 0.69 \\
     \hline
    \end{tabular}
    \end{minipage}
\end{table}
\subsection{Model Training}

To train our LSTM model, we utilize K-Fold cross validation with 10 stratified folds, where each fold preserves the distribution of activity classes present in our dataset $D$. 
Each fold includes a training set $d_{train}$ and testing set $d_{test}$, where $d_{train}$ includes 2,968 samples of activities and $d_{test}$ includes 330 samples of activities. To evaluate each fold, we utilize a validation set $d_{val}$ that is split from $d_{train}$, and includes 594 samples. We leverage early-stopping to train our LSTM model. As a result, our model trains for an average of 25 epochs. We train with a batch size of 64 and our LSTM has a hidden state size of 64. We utilize a sparse categorical cross entropy loss via the Adam optimizer \cite{kingma2014adam} with a 0.001 learning rate.

\subsection {Model Evaluation}
Figure \ref{fig:confusion_matrix} and Table \ref{table:modelperformance} illustrate the performance of our LSTM model across all folds. On average, the LSTM model has an average recall rate of 0.67, precision of 0.73, and F1 score of 0.69 in classifying the 10 ADL. 
Specifically, we notice that the `Bathing' and `Sleep' activity have the highest recall rate, whereas `Eat' has the lowest recall rate. 
We notice that these results correlate with the frequency of each activity class, with the exception of activity class `Other". 
While `Other' is the most frequent activity class in the dataset, it has a lower recall rate compared to `Bathing' and  `Sleep' which are the next most frequent activities. 
We suspect this is due the variability in the `Other' activity itself as it can represent a variety of different activities. 



\begin{table}[t]
\centering
\caption{Example explanations generated by each of the deployed XAI models.}
\label{explanation-examples}
{\small
\begin{tabular}{|>{\centering\arraybackslash}p{1.2cm}|p{3.7cm}|p{3.7cm}|p{3.7cm}|}
\hline
\textbf{Type}  &
  \multicolumn{1}{c|}{\textbf{Activity 1: Leaving Home}} &
  \multicolumn{1}{c|}{\textbf{Activity 2: Cooking}} &
  \multicolumn{1}{c|}{\textbf{Activity 3: Relaxing}} \\ \hline
\textbf{LIME} &
\small  The activity is 'leaving home' because at 14:01 the front door was open and movement near front door was detected, and then 10 minutes later the thermostat near the kitchen read moderate temperatures. &
  \small The activity is `cooking' because at 17:45 the thermostat near the kitchen read high temperatures, 5 minutes later the thermostat near the bathroom read moderate temperatures and then a minute later the thermostat near the bathroom read moderate temperatures. &
 \small  The activity is `relaxing' because at 17:11 the thermostat near the kitchen read moderate temperatures, 10 minutes later the thermostat near the kitchen read high temperatures and then 10 minutes later the thermostat near the bathroom read high temperatures. \\ \hline
\textbf{LIME+} &
\small   The activity is `leaving home' because at 14:01 the thermostat near the kitchen read moderate temperatures for 15 minutes, the front door was open for 6 minutes, and movement near front door was detected for 8 minutes. &
 \small  The activity is `cooking' because at 17:44 the thermostat near the kitchen read high temperatures for 7 minutes, the thermostat near the bathroom read moderate temperatures for 7 minutes, and the pantry door was open for 4 minutes. &
 \small  The activity is `relaxing' because at 17:03 movement in the TV room was detected for 6 minutes, 18 minutes later the thermostat near the kitchen read high temperatures for 2 minutes and then 10 minutes later the thermostat near the bathroom read high temperatures for 19 minutes. \\ \hline
\textbf{Anchors} &
\small   The activity is `leaving home' because at 14:01 the front door was open, 2 minutes later the coat cabinet door was open, and then 12 minutes later the front door was open. &
  \small The activity is `cooking' because at 17:45 movement near the pantry was detected, 4 minutes later movement near the bathroom sink was detected and then a minute later movement in the living room was not detected. &
  \small The activity is `relaxing' because at 17:08 movement on the TV room couch was detected and 24 minutes later movement in the living room was not detected. \\ \hline
\textbf{SHAP} &
\small  The activity is `leaving home' because at 14:01 the front door was open, 13 minutes later movement in the living room was not detected and then a minute later the front door was open. &
 \small  The activity is `cooking' because at 17:33 movement on the TV room couch, 11 minutes later movement at the kitchen entrance was detected and then 6 minutes later movement in the kitchen was detected. &
 \small  The activity is `relaxing' because at 17:23 the thermostat near the kitchen read high temperatures, 9 minutes later movement on the TV room couch was detected and movement in the living room was not detected. \\ \hline
\end{tabular}
}
\vspace*{-1em}
\end{table}

\section{XAI for Activity Recognition}
\label{sec:xai-models}

Our work utilizes XAI methods, namely LIME, SHAP, and Anchors, each of which are model-agnostic and can explain any black-box classifier. 
In our explorations, we utilize these XAI models to generate explanations for the predictions of our LSTM HAR model and consequently explain a predicted activity. These models have been extensively utilized in a wide spread of domains to explain LSTM-based model decision making \cite{thorne2019generating, ribeiro2018anchors, garcia2020shapley, wu2018faithful}.
In order to keep our formulations consistent with the general XAI field, we let $f$ represent a  model and $f(x)$ represents the predicted class of an input $x$. The output of each XAI model, $O$, represents the set of features that best explain $f(x)$. 
Note that a feature in our application represents the triple ($s_{i}$, $v_{i}$, $k_{i}$) as defined in Section \ref{sec:AR-LSTM} for the three XAI models under investigation.
Additionally, $x$ corresponds to $S^{'}$ as defined in Section \ref{sec:overview:problem}. 
Table \ref{explanation-examples} provides example explanations $\mathcal{E}$ for each XAI model. 

In this section, we present technical overviews of the three XAI models and how we utilized them to generate explanations for activity recognition in smart home scenarios.
We also introduce an extension of LIME, LIME+, through which we explore how the temporal nature of explanations affects user perception of the explanation's value.
We utilize the output features of each XAI model to generate associated natural language explanation $\mathcal{E}$ that are understandable by end users. 

 

\subsection{LIME}
To find a set of features belonging to classification input $x$ that best explains the black-box's prediction $f(x)$, the Local Interpretable Model-agnostic Explanations (LIME) algorithm by Ribeiro et al.\  \cite{ribeiro2016should} trains an interpretable surrogate model $g$ to approximate $f(x)$ in the locality of instance $x$. 
Interpretable surrogate models describe the class of models which can be described as inherently interpretable, such as decision trees or linear models \cite{adadi2018peeking}. 
To approximate the decision making of the black-box  model for an input $x$, LIME fits $g$ to a new dataset containing samples $\{z_{1}...z_{n}\}$, which are perturbed from $x$, the instance being explained. 
The result is a trained surrogate model which can provide a local explanation faithful to $f(x)$, but it is not guaranteed to generalize to other predictions of the same class.

Equation \ref{lime_eq} defines LIME's objective function and how the output of LIME, $O_{LIME}$, is derived, where $O_{LIME}$ represents the top $b$ contributing features that can explain $f(x)$: 
\begin{equation}
\label{lime_eq}
{O_{LIME}}(x) = {argmin_{g \in G}} \mathcal{L}(f,g,\pi_x) + \Omega(g)
\end{equation}

Particularly, $\pi_x$ defines the proximity measure, or how close samples $\{z_{1}...z_{n}\}$ are to $x$. $\mathcal{L}(f,g,\pi_x)$ expresses the mean squared error between $f(x)$ and $g(x)$ weighted by $\pi_x$ and $\Omega(g)$ defines the complexity of $g$ (e.g., the depth of a decision tree or the number of non-zero weights of a linear model). 
As such, LIME learns a surrogate model $g$ from a set of surrogate models $G$ that best approximates $f$ in the local region defined by $\pi_x$. 
The outputs $O_{LIME}$ = \{$o_{lime_{1}}, o_{lime_{2}}..o_{lime_{b}}$\} represent the top $b$ features that explain the prediction $f(x)$. 
The top $b$ features are found via Lasso \cite{tibshirani1996regression} with their corresponding contribution weights calculated via Least Squares. 

To generate LIME explanations for smart home activity recognition, we restrict the set of surrogate models $G$ to represent linear models such that $g(z') = w_g \cdot z'$. We ensure interpretability of $O_{LIME}$ by minimizing the model complexity hyperparameter, $\Omega(g)$. In our application, $\Omega(g)$ controls the number of top contributing features, $b$, outputted by $O_{LIME}$; we let $b$ be set to 3 features. The outputs of LIME, $O_{LIME}$, are then mapped to natural language explanations $\mathcal{E}_{LIME}$, as described in Section \ref{sec:gen-exp}. 

\begin{figure}[t]
\centering
\includegraphics[width=0.8\columnwidth]{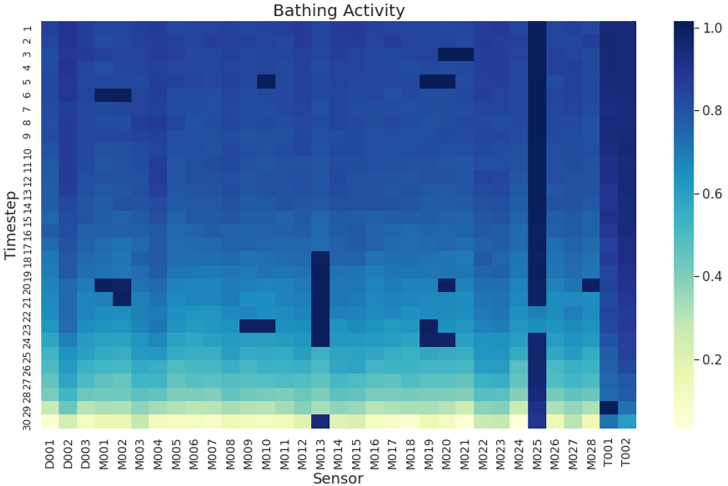}
\caption{Distribution of LIME's feature contributions for an instance of  `Bathing'. Note how M013 (bathroom motion sensor) and M025 (closet motion sensor) show strong contribution to the instance's classification over many consecutive timesteps.}
\label{fig:heatmap}
\vspace*{-1em}
\end{figure}

\subsection {LIME+}
\label{sec:lime+}
From the outputs of $O_{LIME}$, we observe that in instances of longer activities $\mathcal{E}_{LIME}$ identifies features corresponding to the same sensor at multiple consecutive timesteps. 
To further investigate this phenomenon, we analyze the outputs of $O_{LIME}$ across all $M$ features used by the black box model to make a prediction. 
Figure \ref{fig:heatmap} illustrates the relative contributions of all $M$ features for an instance of the activity `bathing'. 
We observe that the activity recognition model recognizes the importance of sensor events \textit{over a period of timesteps} rather than over a single timestep for given activities. 
Specifically, we observe that M013 (bathroom motion sensor) and M025 (closet motion sensor) contribute strongly to the classification `bathing' over many consecutive timesteps which is not captured by $O_{LIME}$.
We posit that it is more unnatural to explain that a sensor event is important towards an activity because of a single timestep rather than a duration of timesteps in the case of longer activities. 
For example, it is more natural to explain that a resident is sleeping because he/she were present in the bedroom over a period of time rather than only because he/she was present at a single moment in time. 

To characterize the duration of a sensor event to an end user, we extend the capabilities of LIME to develop LIME+. 
The feature outputs of $O_{LIME+}$ are represented by ($s_{i}$, $v_{i}$, $k_{i}$, $d_{i}$), which includes an additional duration parameter $d_{i}$ that describes the duration of a sensor event $s_{i}$ that best explains instance $x$. 
Including the durational importance in a LIME+ explanation is motivated by the desire to provide an increased level of context to end users. 
Below we describe how we obtain $d_{i}$. 




We first adapt the original LIME objective function (Equation \ref{lime_eq}) by removing the model complexity parameter, $\Omega(g)$:
\begin{equation}
\label{lime_relaxed_eq}
{O_{LIME+}}(x) = Sort({argmin_{g \in G}} \mathcal{L}(f,g,\pi_x))
\end{equation}

Removing $\Omega(g)$ allows us to understand the relative contributions of \textit{all} $M$ features used by the black box model $f$ to make a prediction $f(x)$ (similar to Figure \ref{fig:heatmap}). 
From this information, we utilize a \textit{Sort} function to sort the outputs of $O_{LIME+}$ by two criteria: sensor contribution as well as sensor uniqueness.

To find $d_{i}$ for a top sensor $s_{i}$ at timestep $k_{i}$, we first calculate the average contribution of $s_{i}$ as well as its standard deviation. 
The number of timesteps consecutive to $k_{i}$ that are within two standard deviations, describe the duration of sensor importance $d_{i}$.
Additionally, to ensure end user interpretability, we only focus on the top three sorted features in $O_{LIME+}$ to generate $E_{LIME+}$ explanations.

\subsection {Anchors}
Anchors by Ribeiro et al.\ \cite {ribeiro2018anchors} explains a prediction $f(x)$ by finding a high-precision rule $R$ called Anchors which represent local, sufficient conditions that explain $f(x)$. 
A rule is composed of a feature and its value. 
The algorithm utilizes a perturbation-based approach to find a set of rules. 
In order to search for a best candidate rule, Anchors formulates a Multi-Arm Bandit problem \cite{kaufmann2013information} commonly utilized in reinforcement learning. 
If there are two or more best candidate Anchors (rules) found, then Anchors outputs the one with the highest coverage.


Formally, Equations \ref{anchor_eq_1} and \ref{anchor_eq_2} define how the outputs $O_{Anchors}$ are calculated.

\begin{equation}
\label{anchor_eq_1}
O_{Anchors} = \max_{R\;s.t.\;P(prec(R)\geq\tau)\geq1-\delta}cov(R)
\end{equation}

\begin{equation}
\label{anchor_eq_2}
cov(R) = E_{D_{(z)}[R(z)]}
\end{equation}

Specifically, $prec(R)$ is the precision of the data satisfying rule $R$, $\tau$ is the precision threshold, $\delta$ is used to guarantee that rule $R$ has at least $\delta$ percent probability of precision above $\tau$, and $cov(R)$ is the coverage of $R$. 
The measure $cov(R)$ represents the percentage of data that satisfied $R$ in a perturbation space $D$. 
If $O_{Anchors}$ includes more than three features to explain $f(x)$, we select the top three features, in order of appearance, to generate natural language explanations for end users.


\subsection {SHAP}
SHapley Additive exPlanations (SHAP) by Lundberg et al.\  \cite{lundberg2017unified} explains a prediction $f(x)$ by calculating Shapley values for all $M$ features used by the black box model. 
The Shapley Value for a feature $j$ represents its marginal contribution to the overall prediction of $f(x)$ \cite{shapley201617}. 
Formally, Equation \ref{shap_eq} defines how the outputs $O_{SHAP}$ are defined. 
Specifically, $M$ represents the number of input features, $x$ represents the input vector, and $\phi_{j}$ represents the Shapley Value for a feature $j$.
\begin{equation}
\label{shap_eq}
O_{SHAP} = \phi_{0} + \sum\limits_{j=1}^M \phi_{j}x_{j}
\end{equation}

While $O_{SHAP}$ outputs a Shapley value for all $M$ features in order to explain $f(x)$, we utilize only the top three features to generate natural language explanations for end users.

\subsection{Explanation Generation}
\label{sec:gen-exp}
Given the output of the four models above, we produce explanations $\mathcal{E}_{LIME}$, $\mathcal{E}_{LIME+}$, $\mathcal{E}_{Anchors}$ and $\mathcal{E}_{SHAP}$ using a template-based approach similar to \cite{elizalde2009generating}. Utilizing domain knowledge, we derive three types of explanation templates ${ET}_{1}$, ${ET}_{2}$ and ${ET}_{3}$ that vary in the amount of included information:
\begin{itemize}
    \item $ET_{1}$: \textit{``The activity is $\langle f(x) \rangle$"}. 
    \item $ET_{2}$: \textit{``The activity is $\langle f(x) \rangle$ because $\langle RT_{1} \rangle$"}
    \item $ET_{3}$: \textit{``The activity is $\langle f(x) \rangle$ because $\langle RT_{2} \rangle$"}
\end{itemize}
${ET}_{1}$ represents a baseline for our work because it simply states the label for the activity, as presented in existing activity recognition systems.  $ET_{2}$ is used to generate $\mathcal{E}_{LIME}$, $\mathcal{E}_{Anchors}$ and $\mathcal{E}_{SHAP}$ explanations, while $ET_{3}$ generates $E_{LIME+}$ explanations via two reasoning templates, $RT_{1}$ and $RT_{2}$, respectively. 

\begin{figure}[t]
\centering
\includegraphics[width=0.89\columnwidth]{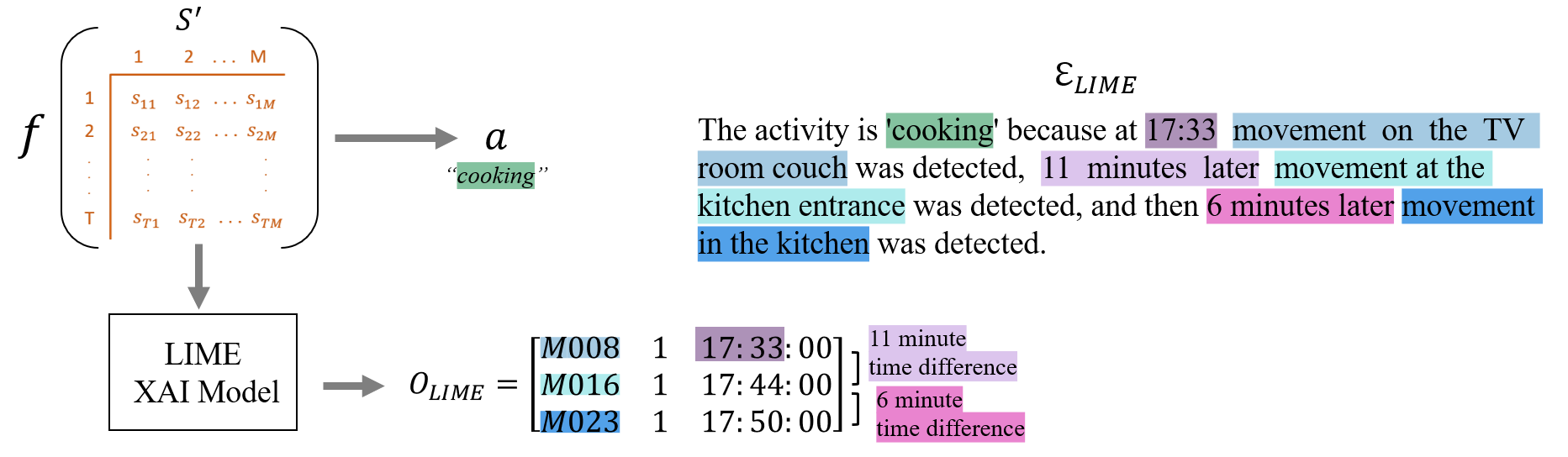}
\caption{Example process for generating $\mathcal{E}_{LIME}$ from $O_{LIME}$ for an activity instance of `cooking', where $f(S^{'})$ is used to generate an activity label $a$, and the LIME XAI model is used to generate $O_{LIME}$. The template style $RT_{1}$ is used to extract the features in $O_{LIME}$ and generate an explanation $E_{LIME}$. }
\label{fig:explanationgeneration}
\vspace*{-1em}
\end{figure}

The following summarizes the overall types of information gathered from both the black-box HAR model as well as from the XAI models: 
\begin{itemize}
    \item The activity prediction of the black-box classifier, $f(x)$ 
    \item Sensor and temporal information ($s_{i}$, $v_{i}$, $k_{i}$) in the features outputs $O_{LIME}$, $O_{LIME+}$, $O_{SHAP}$ and $O_{Anchors}$ 
    \item Duration of a feature importance, in feature outputs $O_{LIME+}$ 
\end{itemize}
Given this information, we define each explanation reasoning template as follows:
\begin{itemize}
    \item $RT_{1}$ as: \textit{\{at $\langle k_{1} \rangle$ $\langle s_{1} \rangle$ $\langle v_{1} \rangle$, $\langle k_{2} - k_{1} \rangle$ minutes later $\langle s_{2} \rangle$ $\langle v_{2}\rangle$, and  $\langle k_{3} - k_{2} \rangle$ minutes later $\langle s_{3} \rangle$ was $\langle v_{3}\rangle$\}}.
    \item $RT_{2}$ as: \textit{\{at $\langle  k_{1} \rangle$ $\langle s_{1} \rangle$ $\langle v_{1} \rangle$ for $\langle d_{1} \rangle$ , $\langle k_{2} - k_{1} \rangle$ minutes later $\langle s_{2} \rangle$ $\langle v_{2}\rangle$ for $\langle d_{2} \rangle$, and  $\langle k_{3} - k_{2} \rangle$ minutes later $\langle s_{3} \rangle$ was $\langle v_{3}\rangle$ for $\langle d_{3} \rangle$\}}.
\end{itemize}
Figure \ref{fig:explanationgeneration} presents an example of $RT_{1}$ being applied in practice to generate $\mathcal{E}_{LIME}$.  Note that in some cases we utilize small variations on these templates to improve the readability and sentence structure of explanations. 

\section{Evaluation: Analysis of XAI Explanations}
\label{sec:xai_exp_analysis}
In this section, we evaluate which XAI method explored in this work (LIME, LIME+, SHAP, Anchors) produces the most sensible explanations for smart home activity recognition models (RQ3).
By sensibility, we refer to how credible the explanations are in explaining the activities modelled by our activity recognition model. Our sensibility metric is similar to prior proposed XAI evaluation metrics that leverage domain experts to measure the ``goodness" of explanations \cite{hoffman2018metrics}. We recruit 
3 ML experts, as volunteers, to classify and evaluate explanations as sensical or nonsensical. We additionally compare and contrast each XAI model's explanations by analyzing the sensors events and their values utilized for each explanation type, as well as report the computational cost of running each model. Our analysis is based on the CASAS Milan scenarios \cite{cook2012casas}.

\subsection{Methodology: Determining Sensical vs Nonsensical Explanations} \label{sec:evaluation:accurate}
\begin{wrapfigure}{R}{2.9in}
    \centering
    \includegraphics[width=2.8in]{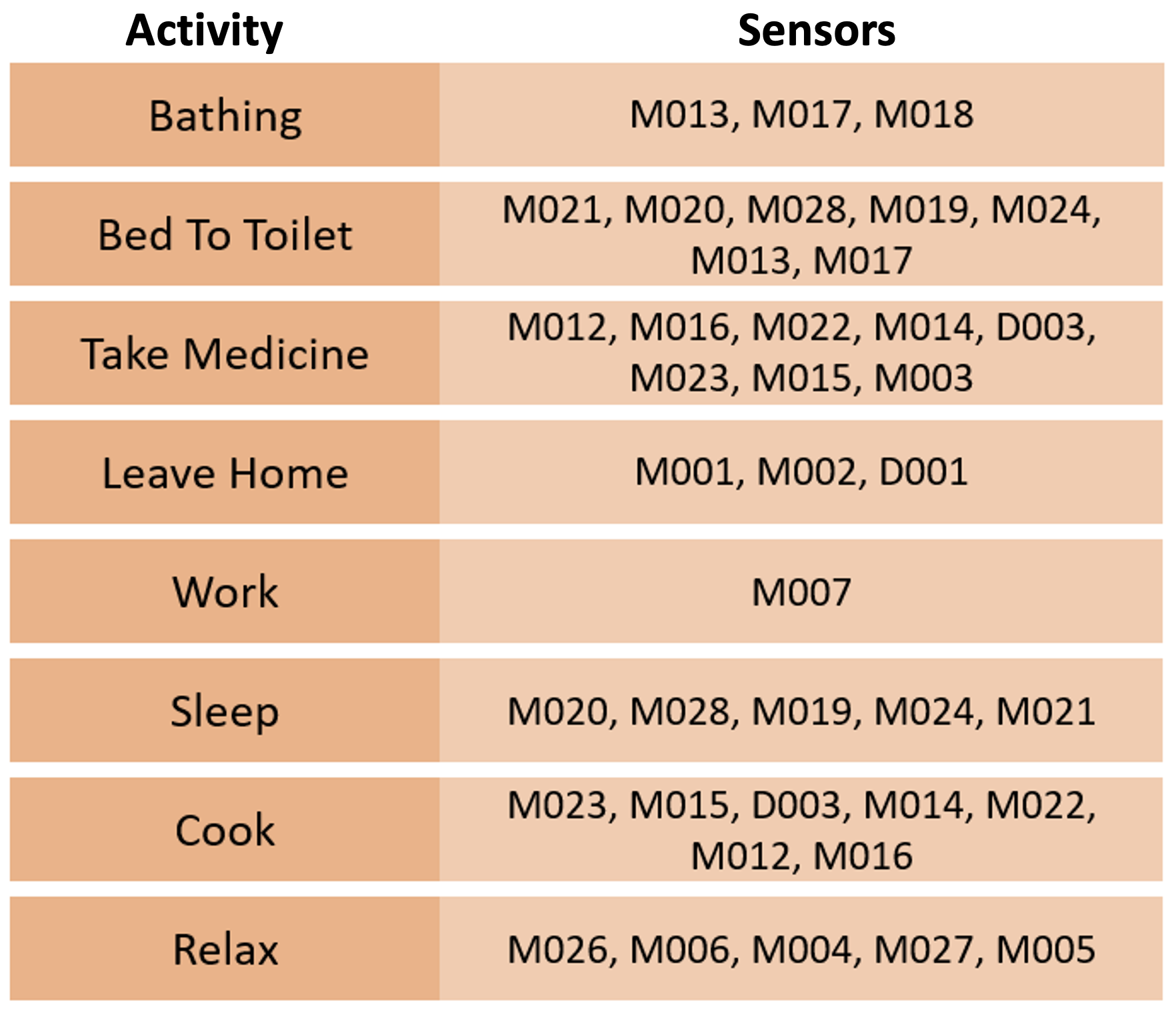}
    \caption{\small Classification set for determining sensical and nonsensical explanations using the Milan CASAS dataset.}
    \label{fig:exp_quality_proc}
\end{wrapfigure}
We leverage a set of classification rules to classify each explanation as ``sensical'' 
or ``nonsensical''.
Note that this is not a perfect classification, and our definition of sensibility does not in itself pertain to whether the explanation fully captures the inner workings of a complex black-box classifier. We define sensibility as whether an explanation is \textit{credible} or can \textit{logically} explain an activity. For example, when explaining the classification of a `Bathing' activity, we would find it credible and logical if sensors near or in the bathroom were referenced.  We would not find it credible if \textit{only} sensors in the kitchen were referenced.  Thus, the mapping between activities and sensors that we find credible constitutes an approximation that we use to measure explanation sensibility for our activity set.

To determine the mapping, we had three ML experts select sensors
whose location in the smart home, extracted from Figure \ref{fig:home_layout}, is proximal to the region of the activity. We validated their selection by performing an inter-rater reliability test, specifically measuring the intraclass correlation coefficient (ICC) \cite{bartko1966intraclass} for each activity. The average ICC was 0.92 with a standard deviation of 0.09. In Appendix \ref{sec:appendix-classifmapping}, we further detail the ICC for each activity.  In Figure \ref{fig:exp_quality_proc} we present the mapping from activities to sensors utilized for the Milan dataset. Table \ref{tab:quality_examples} presents examples of both sensical and nonsensical explanations from all four XAI algorithms.

\subsection{Results}
\label{sec:evaluatation:accurate:results}
\begin{figure}
    \centering
    \includegraphics[width=0.55\textwidth]{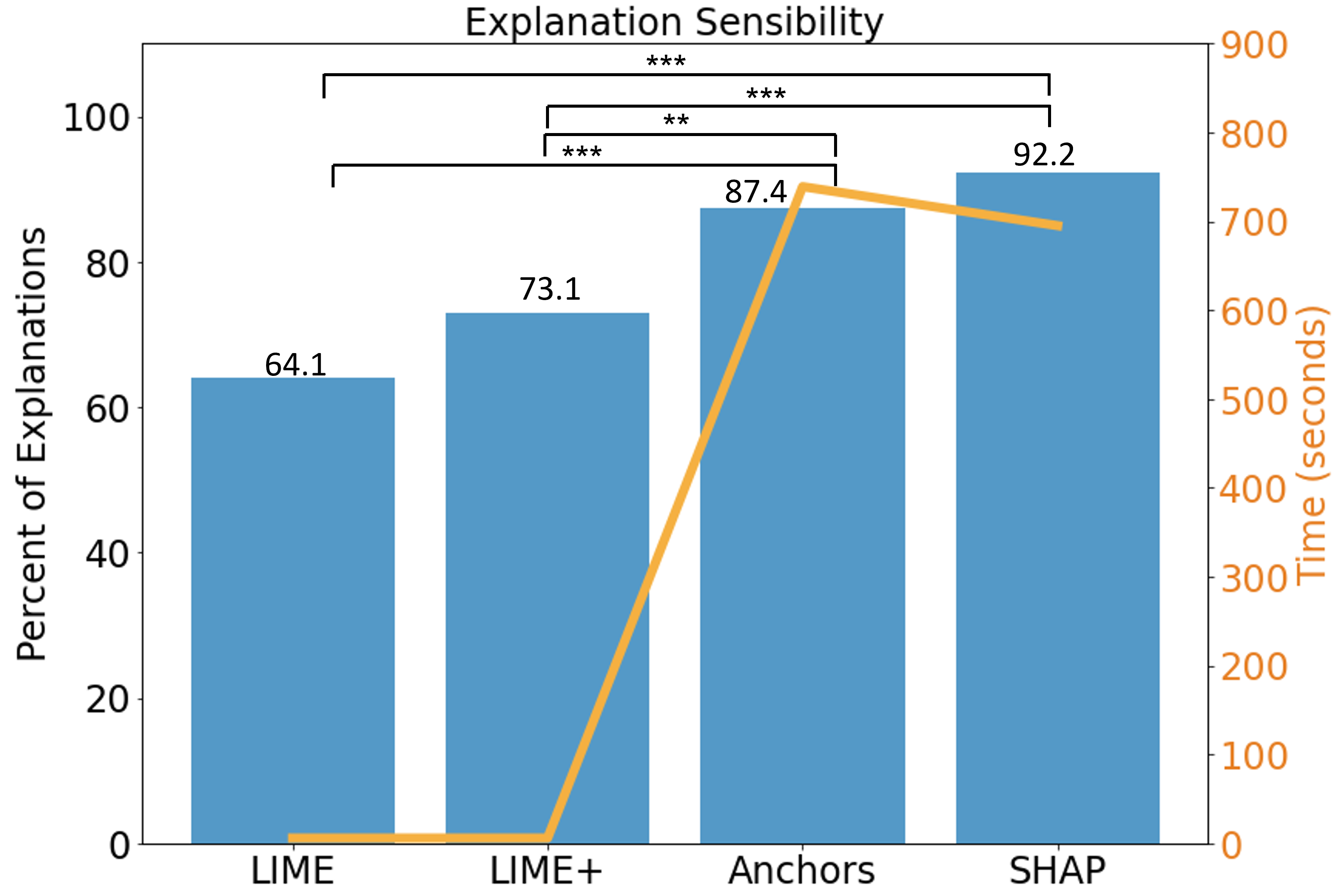} 
    \caption{Blue: percentages of explanations classified as sensible via our classification rules for the CASAS Milan Dataset. Yellow: computational times required to generate an explanation for one activity instance. Statistical significance reported as ** p<0.01, *** p<0.001.}
    \label{fig:acc_cpu}
\end{figure}

To analyze explanation sensibility of each model, we measure the percentage of explanations that are classified as sensible $(ESen\%)$ using the classification rule set introduced above. Note that a total of 167 explanations were classified, and Figure \ref{fig:acc_cpu} illustrates the $(ESen\%)$ for all XAI models (blue) as well as the computation time required for each model to generate an explanation for a single activity instance (yellow). We additionally examine the distribution of sensor types and their values across explanations from each XAI model to understand how these XAI models differ in explanation content, as well as the computational efficiency of each XAI model.

\renewcommand{\arraystretch}{1}
\begin{center}
\begin{table*}[t]
\caption{Example explanations classified as accurate and inaccurate using the classification rules derived for CASAS Milan.}
{\small
\begin{tabular}{|>{\centering\arraybackslash}p{1.1cm}|p{5.8cm}|p{5.8cm}|}
\hline
\multicolumn{1}{|c|}{\textbf{Type}} &\multicolumn{1}{c|}{\textbf{Sensical Explanations}} &\multicolumn{1}{c|}{\textbf{Nonsensical Explanations}} \\
 \hline
 \textbf{$\mathcal{E}_{LIME}$} & 
 The activity is `bathing' because at 15:24 movement in the hallway was detected and then a minute later movement in the shower area was detected. & 
 The activity is `sleep' because between 01:40 - 01:42 the thermostat near the kitchen read low temperatures,  and then 21 minutes later the thermostat near the bathroom read low temperatures
\\ \hline
 \textbf{$\mathcal{E}_{LIME+}$} & 
 The activity is `bathing' because at 15:24 movement in the shower area was detected for 2 minutes, movement in the hallway was detected for 2 minutes, and movement near the bathroom was detected for 2 minutes &
 The activity is `sleep' because at 01:35 the thermostat near the kitchen read low temperatures,  and the coat cabinet door was closed for 27 minutes and then  minutes later the thermostat near the bathroom read low temperatures for 26 minutes
 \\ \hline
 \textbf{$\mathcal{E}_{Anchors}$} & 
 The activity is `bathing' because at 15:24 movement in the shower area was detected, and movement near the front door was not detected & The activity is `sleep' because at 1:35 movement in living room was not detected, 6 minutes later movement near pantry was not detected and 4 minutes later movement in hallway was not detected
 \\ \hline
 \textbf{$\mathcal{E}_{SHAP}$} & 
 The activity is `bathing' because at 15:24 movement in the shower area was detected, and movement near the bathroom sink was detected & 
 The activity is `sleep' because at 01:35 the pantry door was closed, and 37 minutes later movement near the fridge was not detected and the thermostat near the kitchen read low temperatures 
 \\ 
\hline
\end{tabular}
}
\label{tab:quality_examples}
\end{table*}
\vspace*{-1em}
\end{center}

\begin{figure}
    \centering
    \begin{minipage}{0.45\textwidth}
     \centering
    \includegraphics[width=1.0\textwidth]{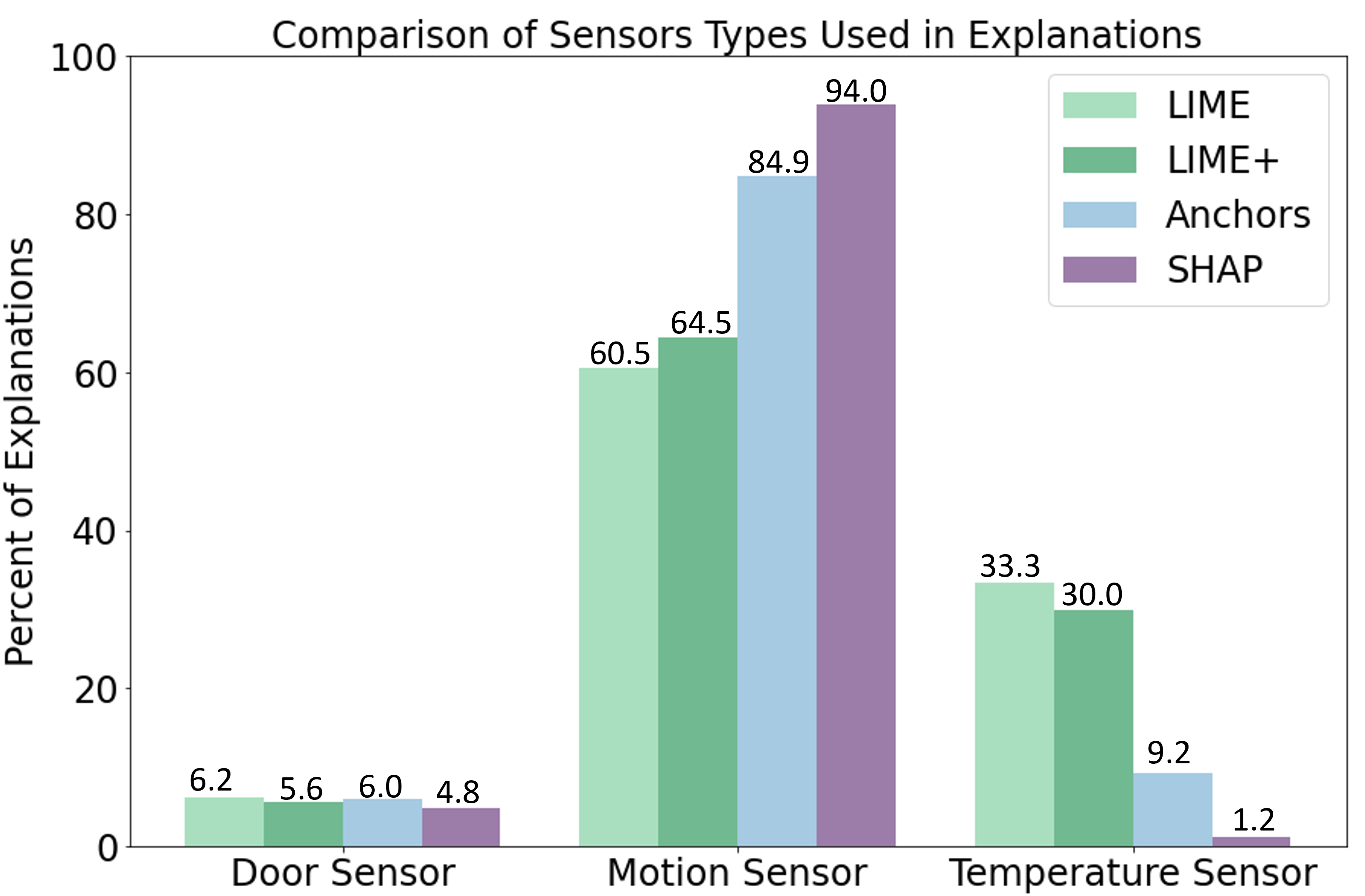} 
    \caption{Percentages of explanations, for each XAI model, that utilize the door sensors, motion sensors and temperature sensors from the CASAS Milan dataset.}
    \label{fig:sensor_types}
    \end{minipage}\hfill
    \begin{minipage}{0.45\textwidth}
     \centering
    \includegraphics[width=1.0\textwidth]{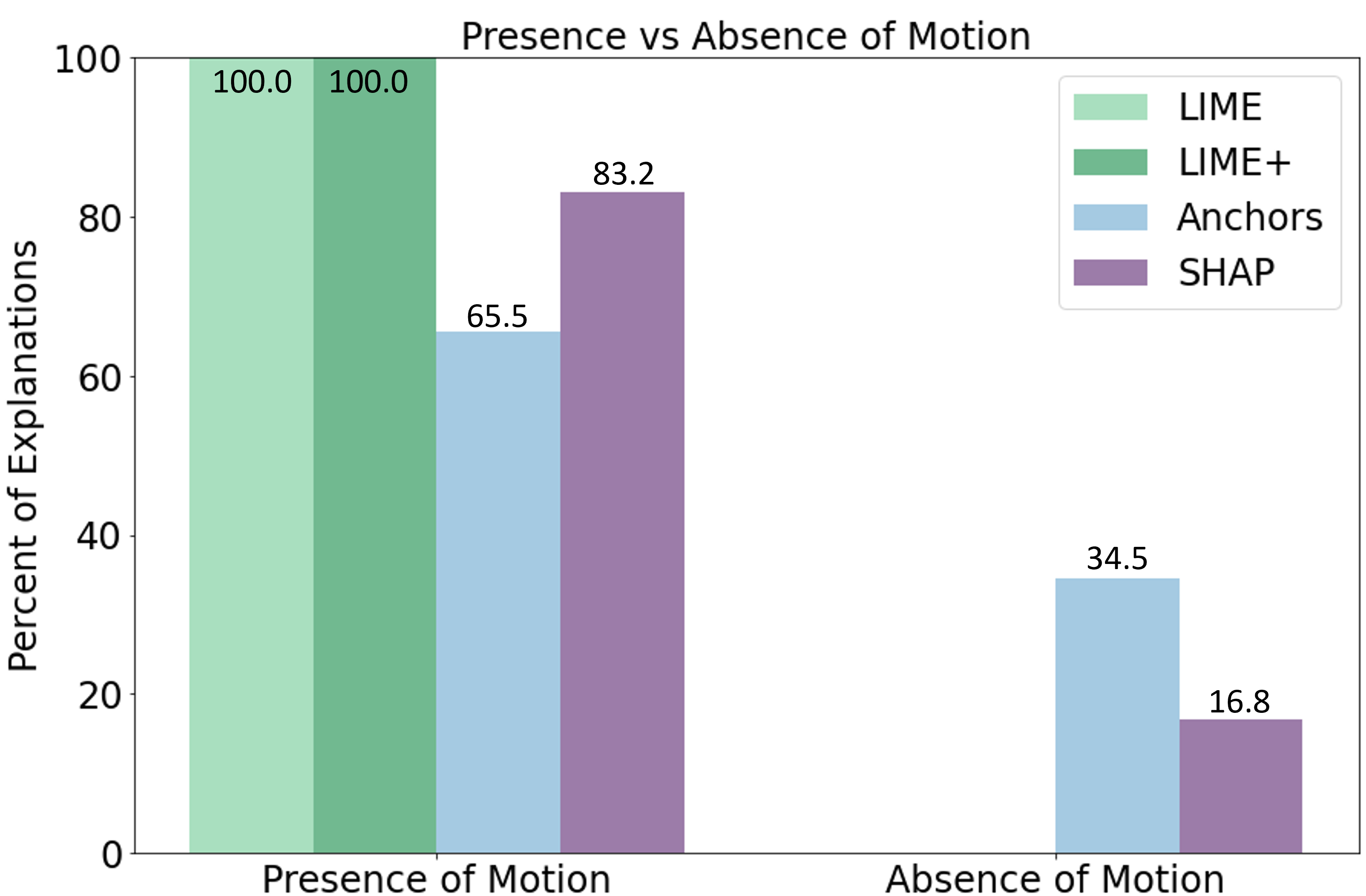} 
    \caption{Percentages of explanations, for each XAI model, that utilize  the absence of motion and presence of motion.}
    \label{fig:absence-presence}
    \end{minipage}
\end{figure}

\subsubsection{Analyzing Sensibility \& Computational Efficiency}
For our analysis, we utilize a one-way Analysis of Variance (ANOVA) with a Tukey post-hoc test. The ANOVA determined a significant difference between the percent of XAI-based explanation accuracies (F(3,664)=18.31, p<0.001). 
We see that $\mathcal{E}_{SHAP}$ have the highest \textit{ESen\%}, while $\mathcal{E}_{LIME}$ have the lowest \textit{ESen\%}. Specifically, we observe that $\mathcal{E}_{SHAP}$ (t(664)=-6.57, p<0.001) and $\mathcal{E}_{Anchors}$ (t(664)=5.45, p<0.001) have a statistically greater \textit{ESen\%} than $\mathcal{E}_{LIME}$. Similarly, $\mathcal{E}_{SHAP}$ (t(664)=-4.48, p<0.001) and $\mathcal{E}_{Anchors}$ (t(664)=3.36, p<0.01) have a statistically greater \textit{ESen\%} than $\mathcal{E}_{LIME+}$. These results show that both SHAP and Anchors generate significantly higher percentages of accurate explanations compared to the LIME and LIME+ XAI models, with the SHAP XAI model generating the most percentages of accurate explanations. 

We also report the computation time required to generate an explanation for one activity instance (yellow in Figure \ref{fig:acc_cpu}) using a quad-core Intel i7-6700K CPU $@$ 4.00GHz with 32GB RAM. We observe that $\mathcal{E}_{LIME}$ and $\mathcal{E}_{LIME+}$ have the shortest computation (5.91 seconds), while $\mathcal{E}_{Anchors}$ explanations have the longest computation (739.1 seconds). We additionally see that $\mathcal{E}_{SHAP}$'s computation time (695.2 seconds) is slightly shorter than Anchors, but similarly longer than $\mathcal{E}_{LIME}$ and $\mathcal{E}_{LIME+}$. While we do not claim superiority of one XAI model over another based on computation time, we demonstrate that with increased accuracy of explanations (\textit{ESen\%}), comes an increased cost in computation time as well. 

\subsubsection{Analyzing Explanation Content} We additionally analyze the types of information presented from each XAI model to understand the similarities and differences between the explanation content by each XAI model. From Figure \ref{fig:sensor_types}, we observe that the motion sensors events (i.e., movement in a particular location of the home), is the most utilized sensor type by each XAI model for describing all activities. However, we see that $\mathcal{E}_{LIME}$ explanations utilize motion sensors less frequently than $\mathcal{E}_{Anchors}$ and $\mathcal{E}_{SHAP}$, while frequently using temperature sensors events (i.e., temperature is at given setting) compared to $\mathcal{E}_{Anchors}$ and $\mathcal{E}_{SHAP}$. This shows that the LIME XAI model tends to focus on the temperature changes as an important contribution towards an activity more than the SHAP and Anchors XAI models. Furthermore, we observe from Figure \ref{fig:absence-presence} that $\mathcal{E}_{Anchors}$ and $\mathcal{E}_{SHAP}$ leverage the absence of movement in a location in addition to the presence of movement, whereas $\mathcal{E}_{LIME}$ and $\mathcal{E}_{LIME+}$ only leverage the presence of movement in a location. These results imply that the SHAP and Anchors XAI models view both the `on' and `off' states of sensors as important factors towards an activity whereas LIME and LIME+ XAI models focus solely on the `on' states.

\section{User Studies: Comparison of XAI Models}
\label{sec:user-eval}
In this section we describe the two IRB-approved user studies through which we examine the usefulness of each explanation type in helping non-AI experts understand smart home activity label outputs. Recall, the user studies consider two different contexts in which remote caregiver monitoring may be applicable.
In the first user study, we consider a remote monitoring smart home scenario in which participants monitor, in real-time, the well-beings of multiple individuals. In the second user study, we consider a remote smart home monitoring scenario in which participants are online and asynchronously review the past-performed activities of a single individual. 

In each user study, we assess five study conditions, four of which are smart home activity labels accompanied with explanation types ( $\mathcal{E}_{LIME}$, $\mathcal{E}_{LIME+}$, $\mathcal{E}_{Anchors}$, $\mathcal{E}_{SHAP}$), and the fifth one is a baseline condition (${Base}$) in which no explanation was provided and only the predicted activity label is presented  (e.g ``The activity is `Bathing'"). The results of our studies contribute to the analysis of RQ1, where we seek to evaluate whether users prefer natural language explanations over simple activity labels from smart home systems, and to RQ2, where we explore whether XAI-model based explanations give users more confidence in the underlying activity recognition model.

\subsection{Study 1: Real-time Remote Smart Home Monitoring}
We first examine the role of explanations in helping remote caregivers monitor the activities of household occupants in a real-time setting. In this study, participants were told that a smart home system known as Timber Smart Home (TSH) was under beta testing and installed in the homes of four occupants (Anna, Margaret, Zeke, Daniel), monitoring the individuals' various activities. Given this information, participants
were presented, via a software tool, with the activity label a household occupant ``performed" along with 1 of the 5 study conditions. For example, when Zeke was ``taking medicine", participants were provided with information that Zeke was ``taking medicine" along with either 1 of 4 explanations ($\mathcal{E}_{LIME}$, $\mathcal{E}_{LIME+}$, $\mathcal{E}_{Anchors}$, $\mathcal{E}_{SHAP}$) or the baseline condition of no additional explanation, $Base$.
Participants were instructed to utilize the provided information and determine whether they could confidently understand the activity performed by the household occupant. Specifically, participants were tasked to answer a number of questions designed to evaluate the effectiveness of the presented study conditions in aiding with remote, real-time monitoring of each household occupants' activities.

Although to participants TSH system was presented as an end-to-end, fully autonomous smart home system, we performed a Wizard of Oz study in which a human (wizard) controlled when activities of each household occupant would be presented to the participant's service tool. \footnote{Wizard of Oz study demonstrated here: https://woz-synchronous.web.app/} Additionally, the four household occupants were presented as real people to more closely emulate the accountability and responsibility of a real, remote caregiver; however, in reality the four individuals were simulated personas. Each simulated household occupant had a designated routine, performing a subset of activities from the CASAS Milan dataset (see Figure \ref{fig:exp_quality_proc}). To help familiarize participants with the TSH system, at the start of the study, participants were provided with a demo video that outlined how the system worked. Specifically, the video showcased example activities being performed by each of the four household occupants as well as the types of questions the participants would be answering.


\textbf{Study Design}: The study was conducted in-person 
and designed as within-subjects given that each participant was exposed to all five study conditions during the study. In order to control for explanation sensibility, and ensure that we are independently analyzing sensibility and explanation type preference, the user study included \textit{only} example explanations that were selected as `sensible' through the process described in Section 6.1. Each participant evaluated a total of 24 activities, across the four household occupants, for which all XAI-models generated sensible explanations. For each activity, a participant was randomly presented with only 1 of the 5 study conditions. Overall, all study conditions for a particular activity were evaluated a total of 3 times, across random participants. At the end of the study, participants were debriefed and told that neither TSH  nor the household occupants were real.


\textbf{Participants}: We recruited 15 participants from our university to serve as remote caregivers. Our participants included 8 males and 7 females, all of whom were 18 years or older (M=22.7, SD=3.5). The task on average took 60 minutes, and participants were compensated with US\$15.00.

\subsection{Study 2: Asynchronous Remote Smart Home Monitoring}
To further examine the role of explanations in aiding remote monitoring, we performed an additional, larger scale user study via Amazon Mechanical Turk (AMT). In this study, participants were presented with the same scenario as in Section \ref{sec:user-pref-study}, in which they played the role of a remote caregiver attempting to monitor the well-being of an elderly patient with Dementia named Pat. The monitoring was not in real-time, but rather \textit{asynchronous}, emulating a scenario in which a caregiver may be reviewing all activities performed by a household occupant at the end of the day. For each activity performed by Pat, participants were provided with Pat's home layout (CASAS Milan home layout), as well all five study conditions. Participants were then asked to respond to a number of questions designed to evaluate the effectiveness of study condition.


\textbf{Study Design}: The study was designed as a within-subjects study, and this time each participant saw and compared all explanation types for each activity. Similar to the synchronous study,  in order to control for explanation sensibility, we only included only example explanations that were selected as ``sensible" through the process described in Section \ref{sec:evaluation:accurate}. Specifically, we evaluated 61 activities for which all XAI-models generated sensible explanations. Given the online setting, and to reduce participant fatigue, each participant on AMT was asked to evaluate at most 8 activities and their explanations; the 61 activities were randomly separated into 8 groups, in which groups 1-7 included explanations for 8 activities, and group 8 included explanations for the remaining 5 activities.\footnote{We performed additional analyses among the groups to ensure grouping bias was not introduced. Please see Appendix \ref{sec:appendix-study2}.} Additionally, to reduce participant bias, the order in which explanations were presented was randomized for each question.

\textbf{Participants}: We recruited 96 individuals from AMT, all of whom had an approval rate greater than 99\%. Specifically,
our participants included 50 males and 46 females, all of whom were 18 years or older (M=38.6, SD =11.6). Each of the 8 groups of activities included 12 participants, thus each activity-explanation pair was reviewed by 12 individuals. Participants in groups 1-7, those who evaluated explanations for 8 activities, took on average 50-60 minutes to complete the study, and were compensated US\$7.00. Participants in group 8, those who evaluated explanations for 5 activities, took on average 20-30 minutes to complete the study, and were compensated with US\$4.00.

\subsection{Metrics}
Prior evaluation methods for explanations have examined qualities such as preference \cite{read1993explanatory}, perceived accuracy \cite{read1993explanatory}, user confidence \cite{ehsan2018learning}, and adequacy in justification \cite{ehsan2018learning}. Thus, across both user studies, we evaluated the effectiveness of each explanation type using similar metrics, detailed below.

\begin{description}
    \item  [Perceived System Accuracy (\textit{SAcc}):] measures participants' self-reported perceptions on the system's ability to identify an activity correctly based on each explanation. Evaluated based on response to the statement ``The smart home correctly identified Pat's activity", measured on a 5-point Likert scale between Strongly Agree and Strongly Disagree (Evaluated in Section \ref{sec:synchronous-study} \& \ref{sec:asynchronous-study}).
    
    \item [Perceived Confidence (\textit{Conf}):] measures the participants' self-reported confidence in the system's tracking abilities. Evaluated based on response to the statement ``Given the explanation, I am confident in the smart home's ability to accurately track Pat's activity", measured on a 5-point Likert scale between Strongly Agree and Strongly Disagree (Evaluated in Section \ref{sec:synchronous-study} \& \ref{sec:asynchronous-study}).
    
    \item  [Perceived Justification Adequacy (\textit{JAdq}):] measures participants' self-reported perceptions on how adequately each explanation justifies a given activity. Evaluated based on response to the statement ``The smart home provides adequate justification as to why Pat is <activity>", measured on a 5-point Likert scale between Strongly Agree and Strongly Disagree (Evaluated in Section \ref{sec:synchronous-study} \& \ref{sec:asynchronous-study}).

    \item [Information Requested (\textit{InR}):] measures percentage of times additional information is requested from an off-site caregiver to validate the household occupant's true activity. Evaluated as a response to ``Given the above explanation, are you confident in the accuracy of TSH system, or would you request additional information from an off-site caregiver?" (Evaluated in Section \ref{sec:synchronous-study}).

    \item [Preference Identification (\textit{PId}):] measures percentage of activities for which a given explanation type was preferred over others.  Evaluated as a response to the question ``Which explanation do you prefer?'' (Evaluated in Section \ref{sec:asynchronous-study})
\end{description}

\begin{figure}
    \centering
    \begin{minipage}{0.45\textwidth}
        \centering
        \includegraphics[width=1.0\textwidth]{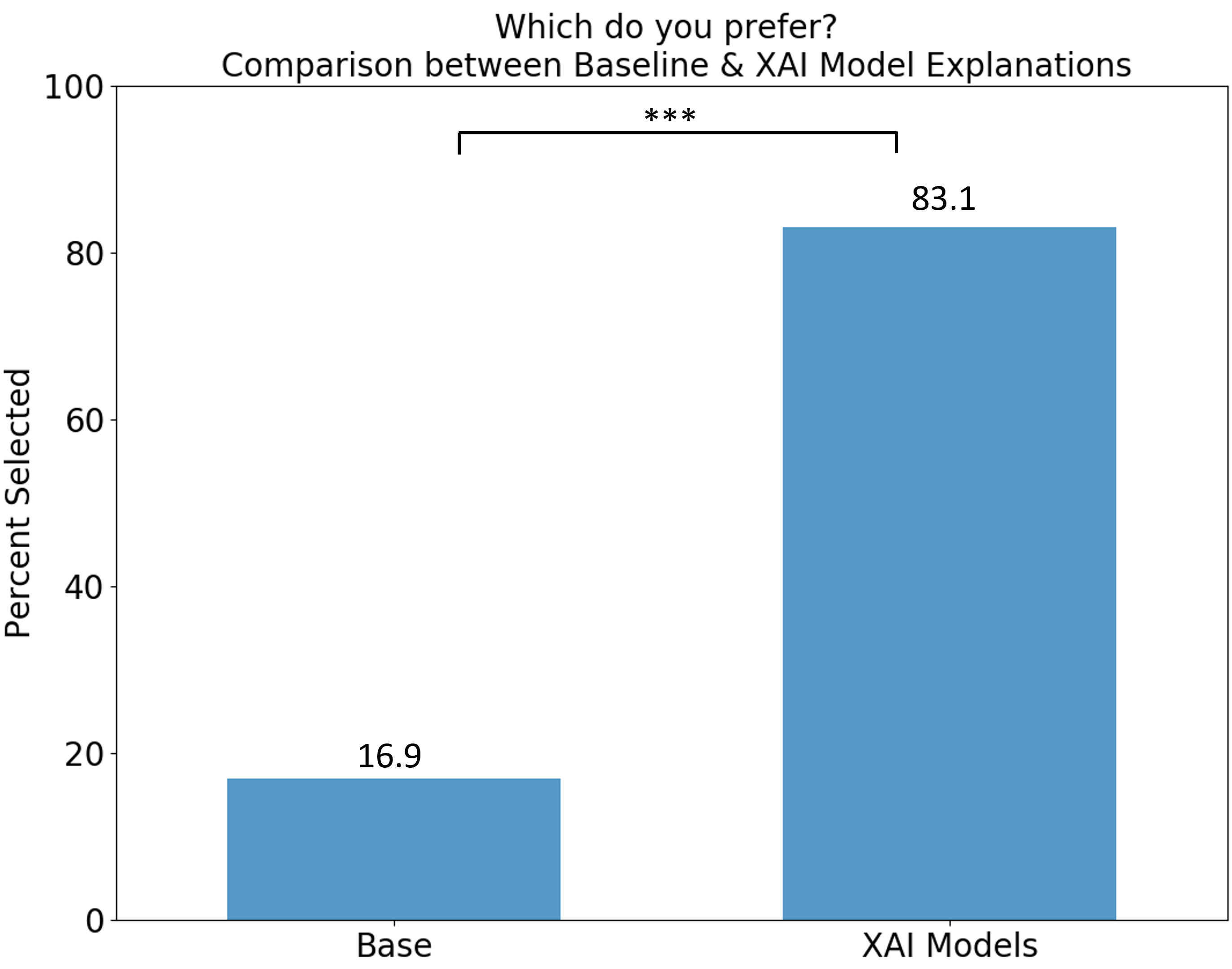} 
        \caption{Comparison of user preferences \textit{(PId)}, showing the preferences between baseline activity labels, ${Base}$, and XAI-Model based explanations. Statistical significance reported as: *** p<0.001.}
        \label{fig:base_xai_prefs}

    \end{minipage}\hfill
    \begin{minipage}{0.45\textwidth}
        \centering
        \includegraphics[width=1.0\textwidth]{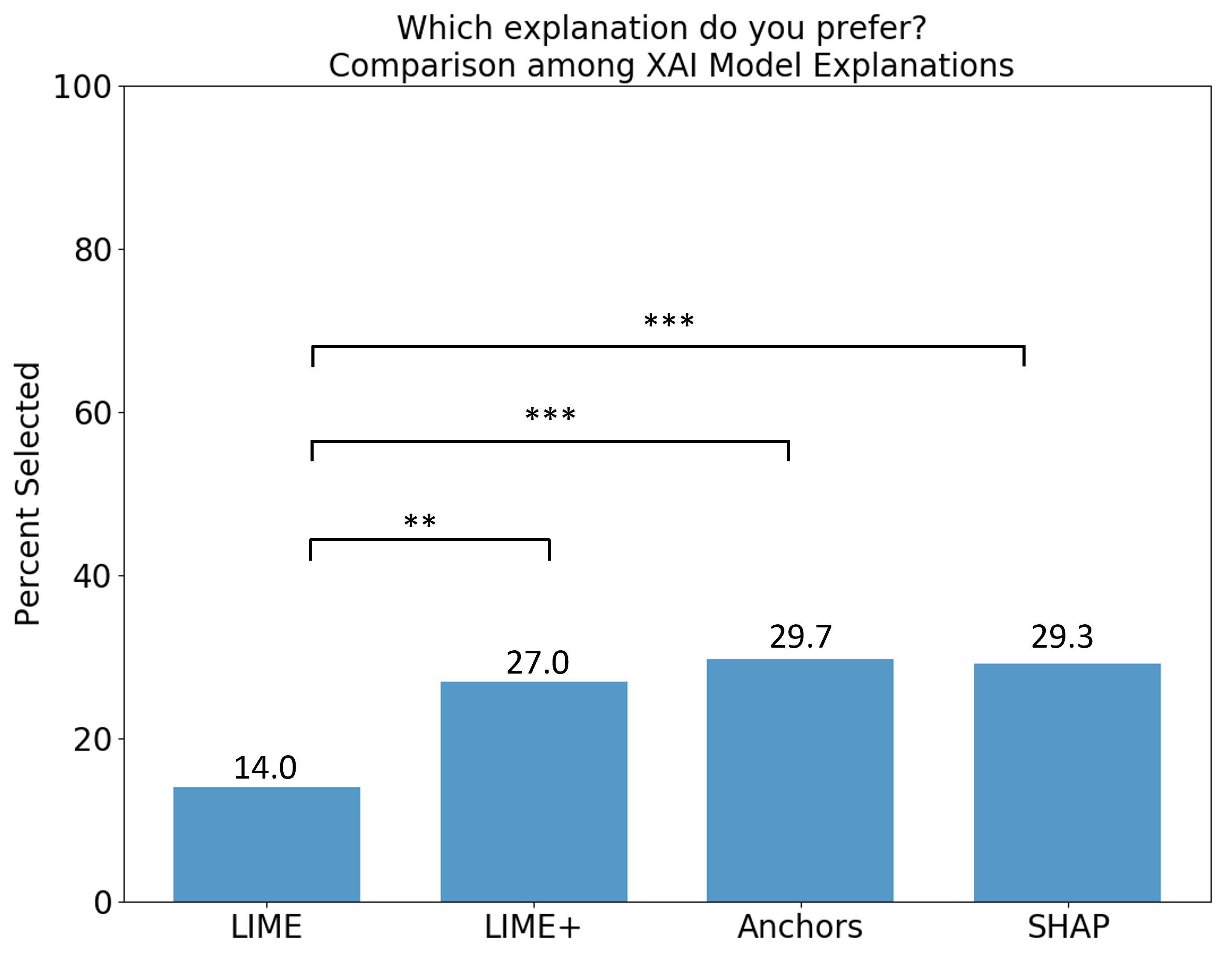} 
        \caption{Comparison of user preferences \textit{(PId)}, showing the preferences among explanations from each XAI model. Statistical significance reported as: ** p<0.01, *** p<0.001.}
        \label{fig:xai_prefs}
    \end{minipage}
\end{figure}

\subsection{Results: User Preference}
\label{sec:RQ1}
One of the first research questions we seek to answer is whether users prefer natural language explanations over simple activity labels (RQ1). While it may seem natural that explanations are preferred, it is an unanswered research question in the community of activity recognition for smart homes. To evaluate our research question, in the asynchronous remote smart home monitoring study, we presented participants with five statements and asked which explanation they preferred as description of Pat's activity. Figures \ref{fig:base_xai_prefs} and \ref{fig:xai_prefs} summarize the results. As seen in Figure \ref{fig:base_xai_prefs}, explanation-based statements (grouped across all conditions) were significantly preferred by participants (83.1\% of tested activities) compared to the baseline, activity-label statements, ${Base}$.   

We further break down the results by explanation model type in Figure \ref{fig:xai_prefs}. We observe participants preferred $\mathcal{E}_{LIME+}$, $\mathcal{E}_{Anchors}$ and $\mathcal{E}_{SHAP}$ explanations over $\mathcal{E}_{LIME}$ explanations, with no strong preference among the top three methods. The distinction between $\mathcal{E}_{LIME}$ and $\mathcal{E}_{LIME+}$ is notable because there are no differences between the two explanation types with respect to which sensory features are being described -- both $\mathcal{E}_{LIME}$ and $\mathcal{E}_{LIME+}$ explanations incorporate the same features.  What differs is that $\mathcal{E}_{LIME+}$ explanations incorporate duration of activities, and LIME explanations do not.  Thus, we conclude that communicating sensor activation duration is found to be more intuitive and valuable by users. Beyond these results, we analyzed our data for correlations between type of activity (e.g., cooking) and explanation type, and we found no correlations between these factors.  As such, we find that explanations generated by these three models perform similarly and are similarly preferred by users.  For developers, this indicates that other factors should be considered when selecting among the three models.  For example, as evaluated in Section \ref{sec:evaluatation:accurate:results}, SHAP produces sensible explanations at a higher rate than other methods.

\textit{Statistical analysis:} We validated the results in Figure \ref{fig:base_xai_prefs} by conducting a Wilcoxon signed-rank test. The results conclude that XAI-model explanations have a significantly higher \textit{PId} compared to ${Base}$ (V = 398.5, $p$ < 0.001). To analyze Figure \ref{fig:xai_prefs}, we utilized the Friedman Test with a post-hoc Nemenyi test to analyze our data given that the data was not normally distributed. The Friedman Test determined a significant difference among the XAI-based explanations ($\chi^2$(3) = 34.6, $p$<0.001). Specifically, we see that $\mathcal{E}_{LIME+}$, $\mathcal{E}_{Anchors}$, and $\mathcal{E}_{SHAP}$ have a significantly higher \textit{PId} compared to $\mathcal{E}_{LIME}$.

\begin{figure}
\centering
\includegraphics[width=0.85\columnwidth]{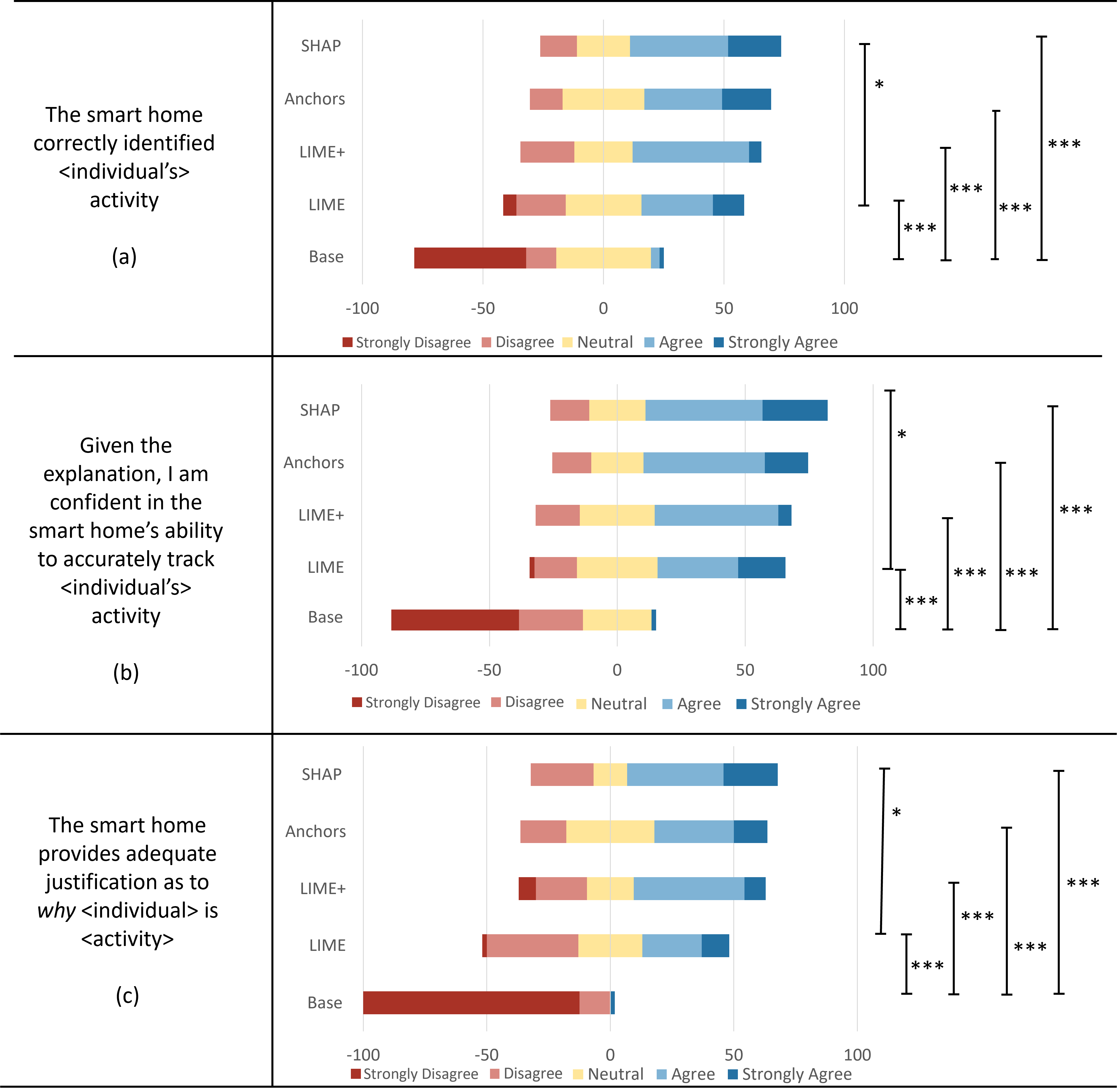}
\caption{Summary of user perceptions for each explanation type in the \textbf{synchronous}, remote smart home monitoring study. Specifically, we examine users' perceived system accuracy \textit{(SAcc)} (a), users' perceived confidence \textit{(Conf)} (b), and users' perceived justification adequacy \textit{(JAdq)} (c). Statistical significance reported as: * p<0.05, ** p<0.01, *** p<0.001.}
\label{fig:statement_data-inPerson}
\end{figure}

\subsection{Results: User Confidence in Activity Recognition Model}
\label{sec:RQ2}
In RQ2 we seek to determine whether explanations give users more confidence in the HAR system. Across both the synchronous and asynchronous monitoring study, we analyze how each explanation type influences participant perception of the smart home's accuracy and capability via three metrics: perceived system accuracy \textit{(SAcc)}, perceived confidence \textit{(Conf)}, and perceived justification adequacy \textit{(JAdq)}. Within the synchronous monitoring study we additionally evaluate perceived confidence via percent information requested \textit{(InR)}. Statistical analysis involving \textit{(SAcc)}, \textit{(Conf)}, \textit{(JAdq)} utilized a one-way repeated ordinal regression with Cumulative Link Mixed Model (CLMM) with a post-hoc Tukey test. \footnote{We validate that the proportional odds assumption required for cumulative link mixed models are met via the nominal and scale tests.} \cite{christensen2018cumulative}. Statistical analysis involving \textit{(InR)} utilized a Friedman's Test with a post-hoc Nemenyi test since the data did not follow a normal distribution.

\begin{figure}
\centering
\includegraphics[width=0.55\columnwidth]{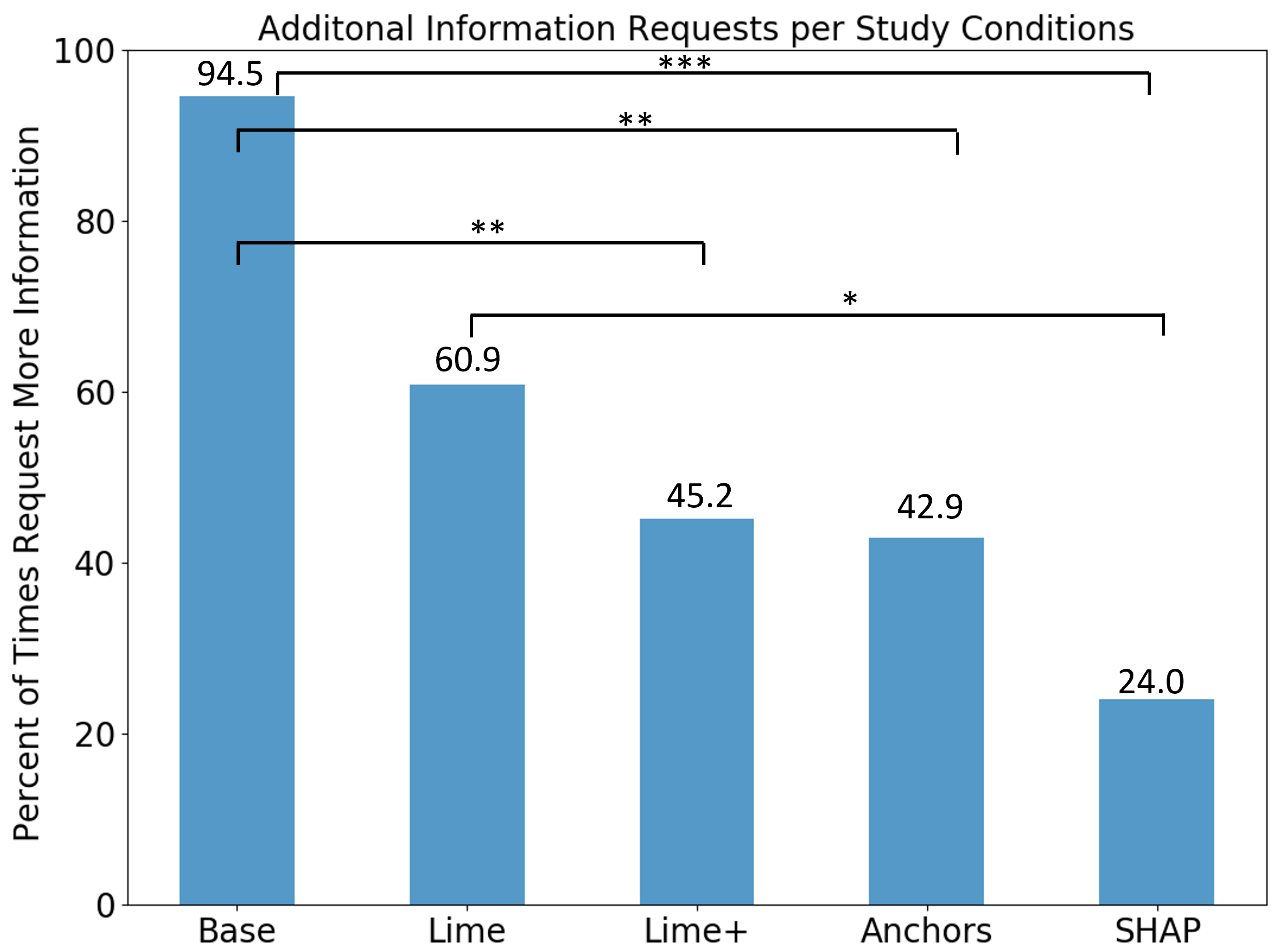}
\caption{Comparison of additional information requested, in the \textbf{synchronous} remote monitoring study, across all explanation types as well as the baseline activity labels condition. Statistical significance reported as: * p<0.05, ** p<0.01, *** p<0.001.}
\label{fig:InR}
\end{figure}

\begin{figure}
\centering
\includegraphics[width=0.85\columnwidth]{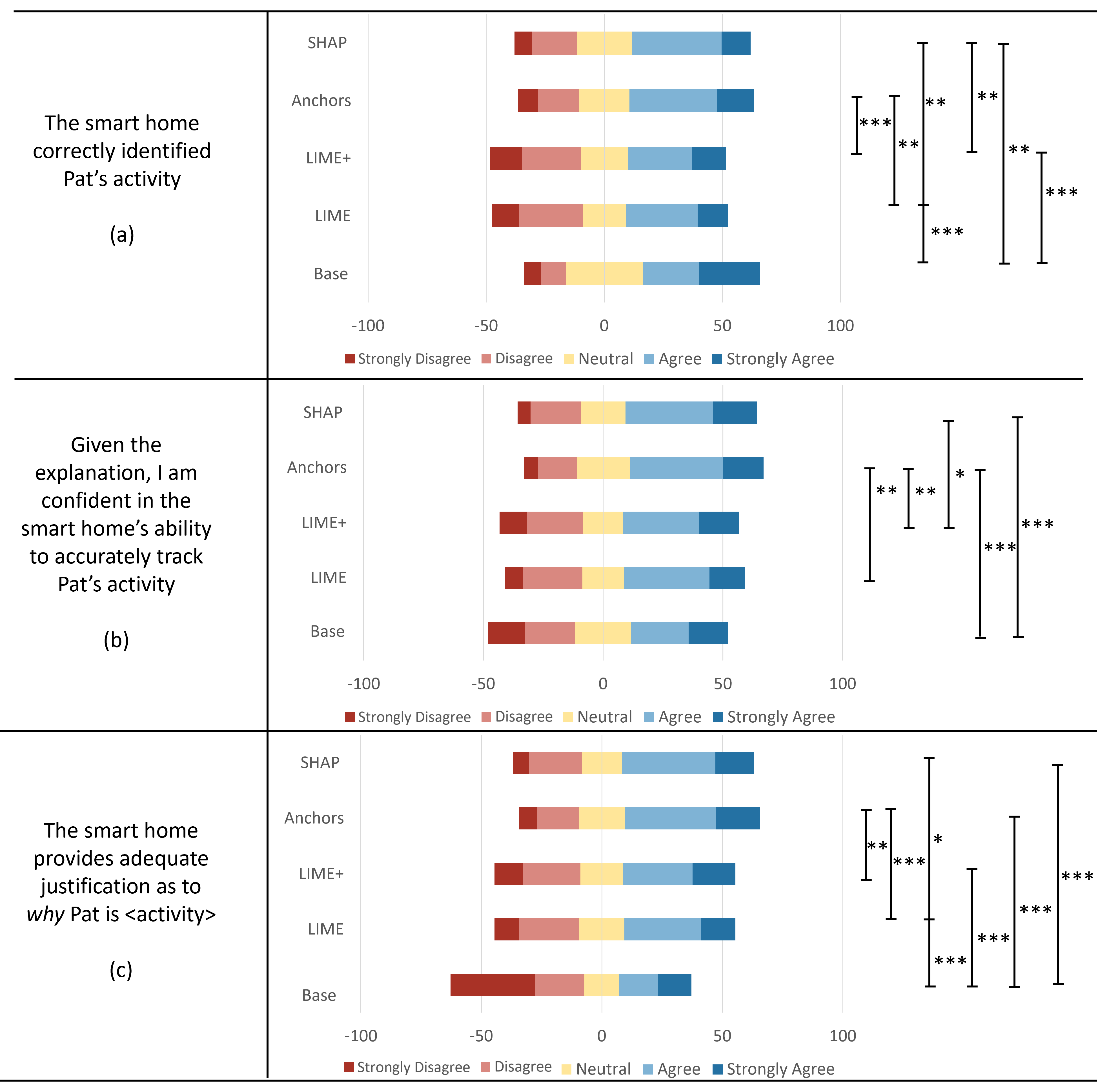}
\caption{Summary of user perceptions for each explanation type in the \textbf{asynchronous}, remote smart home monitoring study. Specifically, we examine users' perceived system accuracy \textit{(SAcc)} (a), users' perceived confidence \textit{(Conf)} (b), and users' perceived justification adequacy \textit{(JAdq)} (c). Statistical significance reported as: * p<0.05, ** p<0.01, *** p<0.001.}
\label{fig:statement_data-AMT}
\end{figure}

\subsubsection{Synchronous Remote Monitoring}
\label{sec:synchronous-study}
Figure \ref{fig:statement_data-inPerson} presents the results of participants' perceptions of each explanation type during the synchronous remote monitoring study. Recall, in this study, participants were monitoring the activities of household individuals in real-time setting. Specifically, Figure\ref{fig:statement_data-inPerson}(a) presents the results for the 
perceived accuracy \textit{(SAcc)} metric, Figure\ref{fig:statement_data-inPerson}(b)
presents the results for the perceived justification adequacy \textit{(JAdq)} metric and 
Figure \ref{fig:statement_data-inPerson}(c) presents the results perceived confidence \textit{(Conf)} metric. Each subfigure presents the statement that participants were asked to agree or disagree with using a 5 point Likert Scale. Across metrics \textit{SAcc}, \textit{JAdq} and \textit{Conf} in Figure \ref{fig:statement_data-inPerson}, we observe a significantly higher rate of ``agrees" across XAI-model based explanations, in comparison to no explanation, ${Base}$. Additionally, we observe the greatest number of ``strongly agrees" towards $\mathcal{E}_{SHAP}$ for each metric. Particularly, with SHAP explanations participants strongly agreed that the smart home correctly identified each individual's activity (\textit{SAcc}), provided adequate justification (\textit{JAdq}) and felt more confident in the HAR model's ability to track activities (\textit{Conf}).

We additionally measured the percent of times participants requested additional information to verify each household occupant's true activity \textit{(InR)}. Evaluating \textit{InR} allows us to understand participants' perceptions on the credibility of each explanation type, serving as another indication of participants' confidence in the HAR system. Figure \ref{fig:InR} presents the results for the \textit{InR} metric. Specifically, we observe that participants have significantly higher \textit{InR} when there is no accompanied explanation in $Base$ and that $\mathcal{E}_{SHAP}$ explanations have the lowest \textit{InR}. Additionally out of the explanation types, $\mathcal{E}_{LIME}$ explanations have the highest \textit{InR}. These results align with those above in that participants found explanations significantly crucial to be confident in the HAR system, and that $\mathcal{E}_{SHAP}$ explanations provided the highest confidence among the explanation types.

\textit{Statistical analysis:} In Figure \ref{fig:statement_data-inPerson}, the one-way repeated ordinal regression with CLMM test determined a significant difference among the study conditions for \textit{SAcc} ($\chi^2$(4) = 111.7, $p$ < 0.001), \textit{Conf} ($\chi^2$(4) = 205.1, $p$ < 0.001), and \textit{JAdq} ($\chi^2$(4) = 245.2, $p$ < 0.001). Specifically we observe that $\mathcal{E}_{SHAP}$ ($p$<0.001), $\mathcal{E}_{Anchors}$ ($p$<0.001), $\mathcal{E}_{LIME}$ ($p$<0.001) , and $\mathcal{E}_{LIME+}$ ($p$<0.001) have significantly higher \textit{SAcc} ratings, \textit{Conf} ratings and \textit{JAdq} ratings compared to  ${Base}$. Moreover, $\mathcal{E}_{SHAP}$ have significantly higher \textit{SAcc} ratings ($p$<0.05), \textit{JAdq} ratings ($p$<0.05), and \textit{Conf} ratings ($p$<0.05) compared to $\mathcal{E}_{LIME}$. In Figure \ref{fig:InR}, the Friedman's Test test determined a significant difference among the study conditions with respect to \textit{InR} ($\chi^2$(4) = 31.9, $p$ < 0.001). Specifically through a post-hoc Nemenyi test we observe that $\mathcal{E}_{SHAP}$ ($p$<0.001), $\mathcal{E}_{Anchors}$ ($p$<0.001) and $\mathcal{E}_{LIME+}$ ($p$<0.001) have significantly lower \textit{InR} in comparison to $Base$. Additionally, we observe $\mathcal{E}_{SHAP}$ has significantly lower \textit{InR} in comparison to $\mathcal{E}_{LIME}$ ($p$< 0.05).

\subsubsection{Asynchronous Remote Monitoring}
\label{sec:asynchronous-study}
Figure \ref{fig:statement_data-AMT} presents the results of participants' perceptions of each explanation type during the asynchronous remote monitoring study. Recall, in this study, participants were not monitoring the activities of individuals in real-time but rather asynchronously as a list of activities previously performed by Pat. Similar to the synchronous study, we measure participants' perceived accuracy \textit{(SAcc)} in Figure \ref{fig:statement_data-AMT}(a), participants' perceived justification adequacy \textit{(JAdq)} in Figure \ref{fig:statement_data-AMT}(b) and participants' perceived confidence \textit{(Conf)} in Figure \ref{fig:statement_data-AMT}(c).

Across the metrics we see trends similar to that in the synchronous study. Out of the explanation types, $\mathcal{E}_{LIME}$ explanations have lower \textit{SAcc}, \textit{JAdq}, and \textit{Conf} ratings than $\mathcal{E}_{SHAP}$ and $\mathcal{E}_{Anchors}$. Additionally participants in the asynchronous study similarly ``strongly disagreed" that $Base$ provided adequate information to justify an individual's activity. Given these results, we see a strong consensus for the need of explainable activity recognition to aid non-expert user understanding of HAR systems and that $\mathcal{E}_{SHAP}$ and $\mathcal{E}_{Anchors}$ explanations lead to high user confidence in HAR systems.

Interestingly, we observe a higher preference for the activity labels with no explanation, ${Base}$, in the asynchronous study which was not the case in the synchronous study. Particularly, ${Base}$ has higher ratings for \textit{SAcc} ratings in comparison to $\mathcal{E}_{LIME}$ and $\mathcal{E}_{LIME+}$. Note that only sensible explanations, classified from Section \ref{sec:xai_exp_analysis}, were given to participants. However, recall from Figure \ref{fig:sensor_types} that $\mathcal{E}_{LIME}$ and $\mathcal{E}_{LIME+}$ most often utilized temperature sensor information as explaining features for activities likely leading to explanations that were consequently perceived as weaker explanations. An example from LIME that supported this trend was: \textit{``The activity is cooking because at 17:45 the thermostat near the kitchen read high temperatures, 5 minutes later the thermostat near the bathroom read moderate temperatures and then a minute later the thermostat near the bathroom read moderate temperatures"}. The results indicate that having read $\mathcal{E}_{LIME}$ and $\mathcal{E}_{LIME+}$ explanations, participants began to doubt that the underlying classification algorithm was working properly, even though it was, preferring $Base$. This trend is strongly demonstrated in the asynchronous study given that participants were exposed to \textit{all} five study conditions for each activity, rating each explanation type as well as $Base$, allowing for a more direct relative comparison between the study conditions for each activity.

\textit{Statistical Analysis}:The one-way repeated ordinal regression with CLMM test determined a significant difference among the study conditions for \textit{SAcc} ($\chi^2$(4) = 71.2, $p$ < 0.001), \textit{Conf} ($\chi^2$(4) = 222.89, $p$ < 0.001), and \textit{JAdq} ($\chi^2$(4) = 49.49, $p$ < 0.001). Specifically, we observe that ${Base}$ ($p$ < 0.001),  $\mathcal{E}_{SHAP}$ ($p$ < 0.01) and $\mathcal{E}_{Anchors}$ ($p$ < 0.01) have a significantly higher \textit{SAcc} rating compared to $\mathcal{E}_{LIME}$. Additionally, we observe that ${Base}$ ($p$ < 0.001),  $\mathcal{E}_{SHAP}$ ($p$ < 0.01) and $\mathcal{E}_{Anchors}$ ($p$ < 0.001) have a significantly higher \textit{SAcc} rating compared to $\mathcal{E}_{LIME+}$. With respect to \textit{Conf}, we observe that $\mathcal{E}_{Anchors}$ ($p$<0.001) and $\mathcal{E}_{SHAP}$ ($p$<0.001) have significantly higher \textit{Conf} values compared to ${Base}$. Additionally, we observe that $\mathcal{E}_{Anchors}$ ($p$<0.01) and $\mathcal{E}_{SHAP}$ ($p$ < 0.05) have significantly higher \textit{Conf} values compared to $\mathcal{E}_{LIME+}$. Finally, with respect to \textit{JAdq}, we observe that $\mathcal{E}_{SHAP}$ ($p$<0.001), $\mathcal{E}_{LIME}$ ($p$<0.001), $\mathcal{E}_{LIME+}$ ($p$<0.001) and $\mathcal{E}_{Anchors}$ ($p$<0.001) have significantly higher \textit{JAdq} ratings compared to $\mathcal{E}_{Base}$. Additionally, we observe that $\mathcal{E}_{SHAP}$ ($p$<0.05) and $\mathcal{E}_{Anchors}$ ($p$<0.01) have significantly higher \textit{JAdq} ratings compared to $\mathcal{E}_{LIME+}$.

\section{Discussion}
\label{sec:discussion}
In this work, we conducted three evaluations of the explainable activity recognition framework we have introduced. Recall our framework leverages leading XAI-model based explanation techniques and a state-of-the-art smart home HAR model to generate natural language explanations that help non-AI experts understand smart home monitoring and decision making. Specifically, the analysis in Section \ref{sec:xai_exp_analysis} answers RQ3 which investigates, from an expert analysis, which XAI methods produce the most sensible explanations and our user studies in Section \ref{sec:user-eval} answer RQ1 and RQ2 which investigate, from a non-AI expert perspective, which XAI-model based explanation is most preferred (RQ1) and do such XAI-model based explanations provide users with more confidence in HAR model behaviors (RQ2). Given our analyses, our work demonstrates that an explainable activity recognition framework, like the one proposed in our work, is crucial for helping users understand and gain confidence in HAR system behavior (Section \ref{sec:synchronous-study} \& Section \ref{sec:asynchronous-study}) as well as produce natural language explanations that are accessible and preferable (Section \ref{sec:RQ1}).

Given these findings, one question that remains is \textbf{which XAI-model based method is ultimately most effective for producing explainable activity recognition?} To answer this question, we consider factors across all our evaluations, including model sensibility, computational efficiency and user preference and confidence. Across our evaluations, we found that: 

\begin{itemize}
    \item SHAP and Anchors generated the highest percentage of sensible explanations (92.2\% and 87.4\% compared to 64.1\% of LIME), at the cost of  higher computational resources (Section \ref{sec:evaluatation:accurate:results});
    
    \item LIME was the least preferred explanation method among non-AI experts in a direct comparison (Section \ref{sec:RQ1});
    
    \item LIME and LIME+ explanations both caused more users to doubt the correctness of the activity recognition model (Section \ref{sec:asynchronous-study}, but that LIME explanations, particularly, evoked the need for additional information a significant amount of times (Section \ref{sec:synchronous-study});
    
    \item SHAP significantly increased user confidence in the accuracy of the HAR model and smart home system across both asynchronous and synchronous remote care-giving scenarios (Section \ref{sec:asynchronous-study} \& Section \ref{sec:asynchronous-study}); and
    
    \item SHAP explanations required considerably less requests of additional information from  off-site household occupants during real-time caregiver monitoring in comparison to other XAI-model explanations (Section \ref{sec:synchronous-study}).
\end{itemize}

Taking these factors into consideration, across both the expert and non-expert analysis, we find that the SHAP model results in the most effective explanations. While there is a strong desire for explanations, especially in real-time remote caregiver monitoring, we note that in many instances users did not show a significant preference between the various explanations when asked to indicate preference (Figure \ref{fig:xai_prefs}), suggesting that possibly a broad spectrum of acceptable explanations may exist, including explanations even more effective than the ones producible by current XAI-model based techniques. More generally, these findings lead to new research questions in a number of areas:

\smallskip
\noindent \textbf{The benefits of explanations:}  Our work clearly demonstrates that providing explanations to end users of activity recognition systems is of great value within the domain of smart home automation. As in other areas of XAI, HAR explanations serve to elucidate the decision making process of a complex computational model and enable users to gain understanding of which factors most significantly influenced model outcome \cite{dovsilovic2018explainable}.  Increased user understanding in turn leads to increased confidence in the system.  This finding raises many questions for future work.  For example, how can we improve computational methods for XAI, particularly as they apply to temporal decision making settings? Furthermore, does incorporating XAI into deployed systems smart home lead to improved user satisfaction, trust and long-term use?

\noindent \textbf{The negative effects of explanations:} We observed that explanations can in some cases have a negative effect on user perception of the system.  Specifically, we saw that confusing explanations led some participants to doubt the accuracy of the system's activity recognition output even when it was correct. These findings are similar to prior work that demonstrate the negative effects of explanations negatively affecting user perceptions even under correct classifications \cite{lim2011investigating}. In our work we specifically see that an unsuitable explanation can be worse than no explanation at all. Today's XAI models are not designed to reason about the quality of their explanation.  An important direction for future work is to consider how XAI models can communicate confidence in how well they are able to explain a particular input.

\noindent \textbf{The content of an explanation:} In this work we studied different computational techniques for identifying the underlying state features that contribute to an HAR classification.  Given such features, we used a template approach for generating explanations that sought to present those features in an interpretable way.  Thus, we studied what information \textit{could} be encoded in an explanation, but did not exhaustively study what information \textit{should} be included.  In other words, we did not try to find the ``perfect'' explanation (if one exists) for activity recognition in smart homes.  Thus, an open question remains of what information would be ideal to include, such that future computational XAI research can be guided toward techniques that extract such information. Additionally, we observe from our analyses that all XAI models focus on slightly differing sensor features for generating explanations, making it difficult to objectively evaluate the true accuracy of each XAI model. Addressing the ``disagreement" problem among XAI methods is an ongoing challenge within the field of XAI \cite{krishna2022disagreement}. Improving the effectiveness of XAI methods, themselves, can lend to improved explainable activity recognition systems.

\noindent \textbf{Differences in user preferences:} Our results indicate that users have a broad range of opinions relating to any given explanation, suggesting that personal preference may play a significant role. Existing XAI methods typically provide one type of explanation for all users.  An interesting question for future work is to consider how explanations could be customized, with respect to length, type, or level of detail, for individual users.

\noindent \textbf{Longitudinal effects of XAI systems:} Due to the relative novelty of XAI systems, and their absence in prior activity recognition systems, we have no findings regarding the long-term effects that providing explanations will have on users.  While it is encouraging to think that explainability will lead to greater trust in a system, it is also possible that users will find explanations boring, distracting or annoying.  

\noindent \textbf{Implications of Real-World Use:} Through our user studies, we observe the benefits of SHAP-based explanations in scenarios of real-time and asynchronous remote caregiver monitoring. However, since participants in our study are asked to play the role of a caregiver, an open question that remains is -- how useful are these explanation techniques to real, existing caregivers? Analyzing the efficacy of our explainable activity recognition framework with real-world caregivers in future user studies can identify methods to improve the robustness of such explanations in real-world applications.

\noindent \textbf{Implications for Activity Recognition:} An important research question that remains unanswered is how activity recognition systems can best facilitate explainability -- are there changes that can be made to the way features are represented or models are learned that would improve the transparency of HAR models?  Or, alternatively, can XAI lead to improvements in activity recognition itself?  Is it possible to leverage the understanding that users gain through XAI to enable users to train, teach, or instruct HAR systems (through active learning or demonstrations) in order to further customize smart home systems and improve performance?

These, and many other research questions remain open at the intersection of activity recognition and explainable AI research.  This paper serves to establish the foundation that we hope will lead to many further advances in this research area.






\section{Conclusion}
\label{sec:conclusion}
While smart home systems have the potential to provide services that can help improve the quality of life of their occupants, they are not perfect systems. Inconsistencies in their underlying activity recognition systems do exist and may cause smart home users (i.e., household occupants, remote caregivers) to wonder "Why did the smart home do that?"
In this work, we contribute an explainable HAR framework that leverages explainable AI techniques and a leading HAR model to generate meaningful explanations to end users about a smart home's activity recognition. We first preserve the multivariate nature of smart home sensor data (both timestep and sensor event information) and then leverage state-of-the-art XAI models--LIME, SHAP, and Anchors--to extract the most contributing feature information towards an activity recognition label.  In our framework, we utilize the outputs of each XAI model to generate template-based natural language explanations that explain the HAR model's predictions. We evaluate our explainable HAR framework through two evaluations. First, we perform an expert analysis and examine the explanation sensibility as well as their computational efficiency of all XAI models to understand how transferable these XAI techniques are to explaining multivariate activity recognition data. Second, we evaluate the XAI explanation types with end users, in the context of synchronous and asynchronous remote monitoring, to understand user preferences of each explanation type as well as perceived confidence in the HAR system. From our results, we find that users significantly prefer explanations from our explainable activity recognition framework, underscoring the need for such systems. Additionally, we find SHAP based explanations produce the most sensible explanations compared to the other XAI models. 
We also observe that SHAP based explanations are most effective in increasing users' confidence in the correctness of an HAR model in both user studies.

\section{Acknowledgements}
This work is in part supported by KDDI Research, and the NSF Graduate Research Fellowship under Grant No. DGE-1650044.

\bibliographystyle{ACM-Reference-Format}
\bibliography{references}


\appendix
\newpage
\section{User Study Grouping Analyses - Section 4}
\label{sec:appendix-study1}

\begin{figure}[h]
\centering
\begin{subfigure}[b]{0.45\textwidth}
    \includegraphics[width=\textwidth]{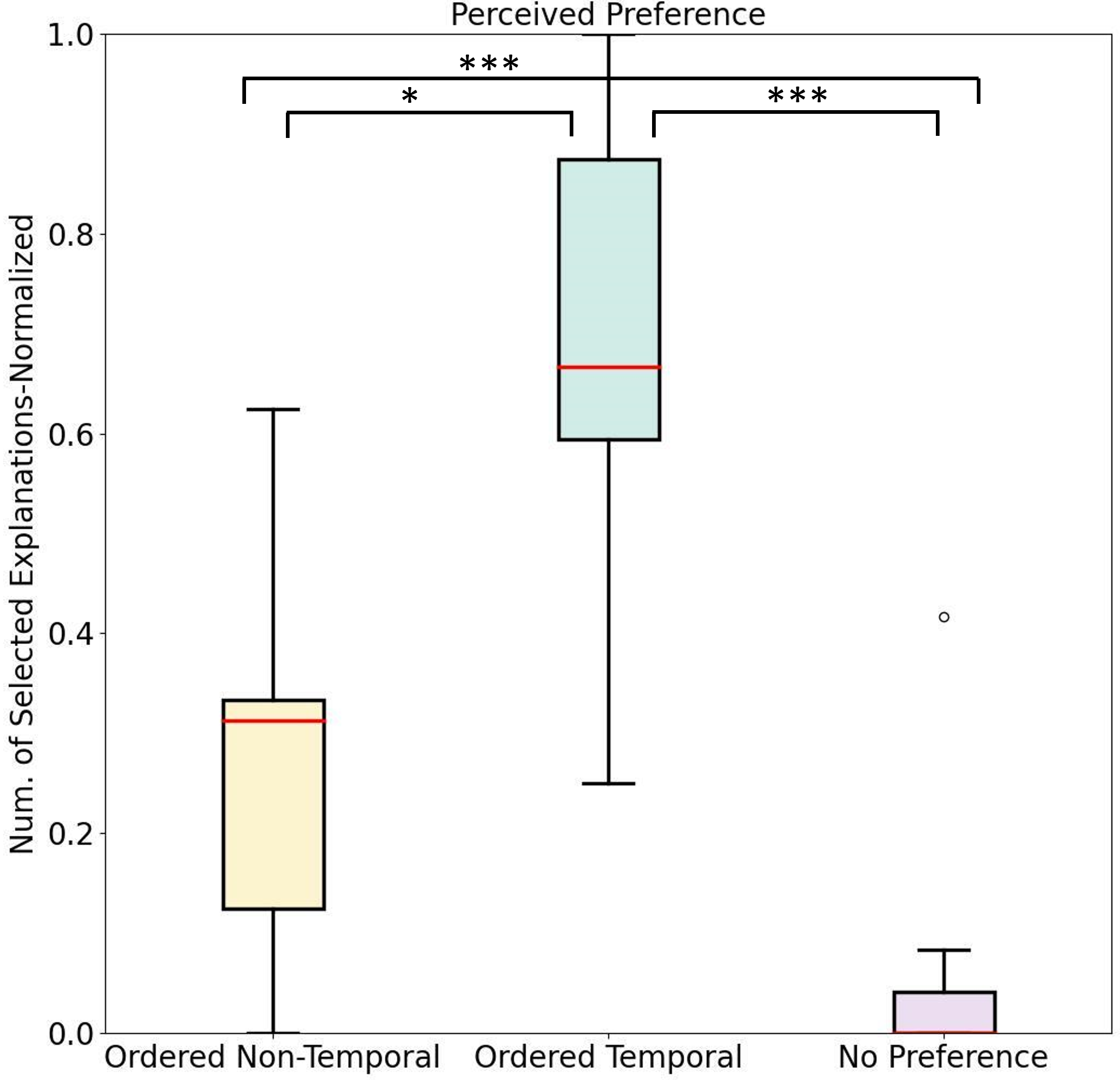}
    \caption{}
    \label{fig:first}
\end{subfigure}
\hfill
\begin{subfigure}[b]{0.45\textwidth}
    \includegraphics[width=\textwidth]{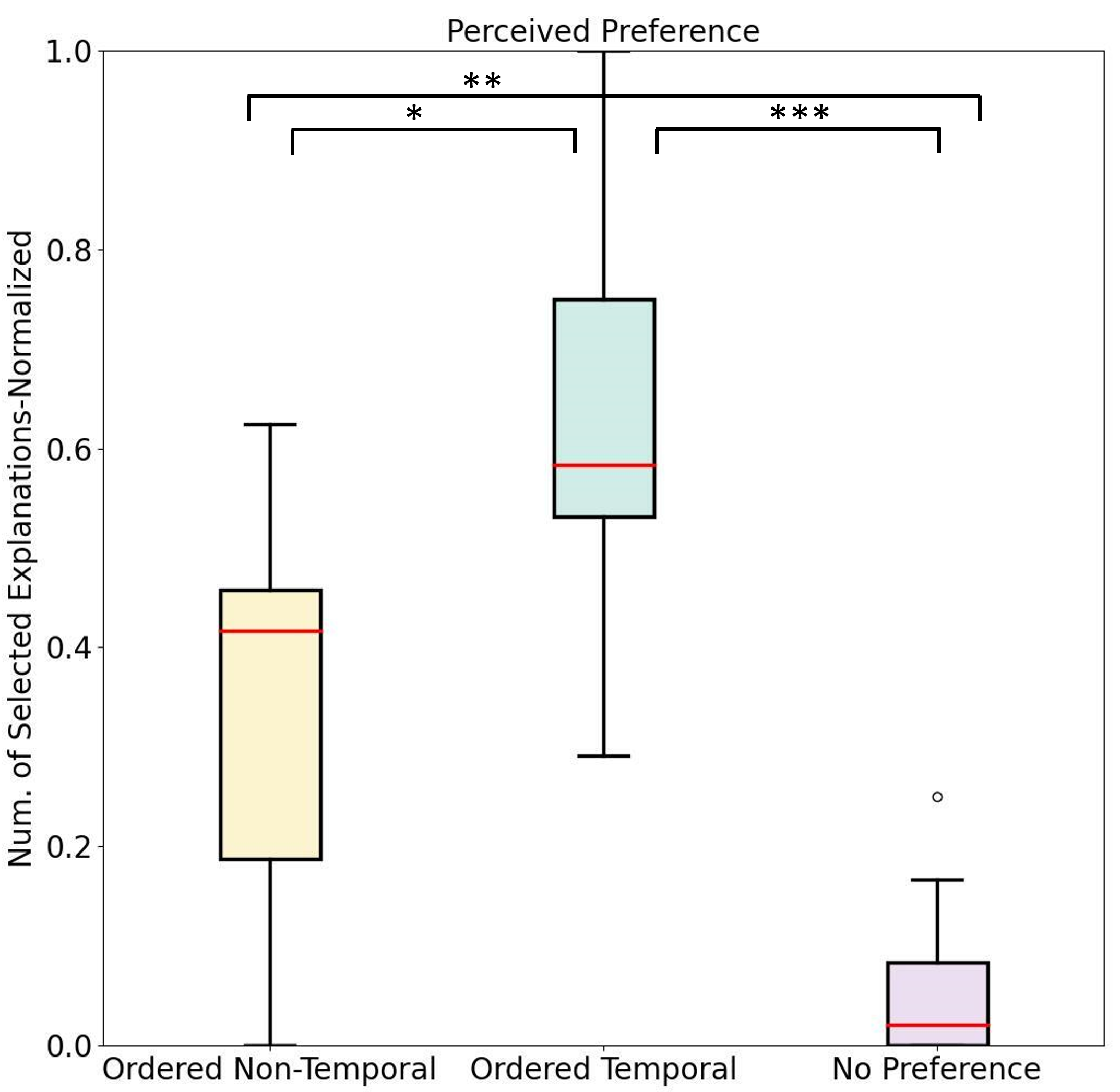}
    \caption{}
    \label{fig:second}
\end{subfigure}
\hfill
\caption{Comparison of user selected preferences towards each explanation type by participant group in which (a) shows the first group's responses, and (b) shows the second group's responses. Significance reported as * p<0.05, ** p<0.01, *** p< 0.001.}
\label{fig:appendix-study1}
\end{figure}

To ensure that grouping bias was not introduced in the study design, we first analyze each of the 2 groups' responses separately. Figure \ref{fig:appendix-study1}(a) and \ref{fig:appendix-study1}(b) show each group's responses. We perform a Friedman's Test and observe statistical significance in both the first group ($\chi^2$(2) = 38.8, $p$ < 0.001) as well as the second group ($\chi^2$(2) = 36.0, $p$ < 0.001). We additionally perform a post hoc Nemenyi test on each group and observe that in both groups participants significantly prefer ordered temporal explanations in comparison to ordered non-temporal explanations ($p$ < 0.05). Given both groups of responses show similar statistical significance, we infer that there is no grouping bias observed.

\section{User Study Grouping Analyses - Section 7.2}
\label{sec:appendix-study2}
We perform a series of statistical analyses to ensure that our study design with grouped participants did not introduce grouping bias. Specifically, we perform a one-way repeated ordinal regression with cumulative link mixed model (CLMM) \cite{christensen2018cumulative} with a post-hoc Tukey test to measure any effect the 8 different groups of participants (independent variable) may have on the Likert responses (dependent variable) for the three statements evaluated in the user study in Section 6. As a summary, from all the statistical analyses, we observe no significant differences between any of the 8 groups of participants, indicating low chances of a grouping bias. Below we detail statistical results for each statement evaluated.

\subsection{Evaluation for Statement in Figure \ref{fig:statement_data-AMT}(a)}
As a recap, Figure \ref{fig:statement_data-AMT}(a) evaluates the Likert responses to the following statement: ``The smart home correctly identified Pat's activity". When conducting the one-way repeated ordinal regression with CLMM, we observe no significant effect of the 8 groups on the Likert responses ($\chi^2$(7) = 9.29, $p$=0.23). For sanity measures, we also report the pairwise comparison results from the Tukey test in Table \ref{tab:proof14a}, and observe no statistical differences between any pairs of groups.

\renewcommand{\arraystretch}{1}
\begin{center}
\begin{table*}[h]
\caption{Tukey Post-Hoc Analysis between Participant Groups for Figure\ref{fig:statement_data-AMT}(a)}
{\small
\begin{tabular}{|>{\centering\arraybackslash}p{1.5cm}|p{2.2cm}|p{2.2cm}|}
\hline
\multicolumn{1}{|c|}{\textbf{Pairwise Groups}} &\multicolumn{1}{c|}{\textbf{Z-ratio}} &\multicolumn{1}{c|}{\textbf{P-Value}} \\
 \hline
 G1 - G5 & 
 -0.225 & 1.000
\\ \hline
 G1 - G6 & 
 -0.314 & 1.000
\\ \hline
 G1 - G7 & 
 -0.186 & 1.000
\\ \hline
 G1 - G2 & 
 0.000 & 1.00
\\ \hline
 G1 - G3 & 
 1.694 & 0.692
\\ \hline
 G1 - G4 & 
 -0.843 & 0.991
\\ \hline
 G1 - G8 & 
 -0.932 & 0.983
\\ \hline
 G5 - G6 & 
 -0.082 & 1.000
\\ \hline
 G5 - G7 & 
 0.020 & 1.000
\\ \hline
 G5 - G2 & 
 0.225 & 1.000
\\ \hline
 G5 - G3 & 
 1.860 & 0.578
\\ \hline
 G5 - G4 & 
 -0.598 & 0.998
\\ \hline
 G5 - G8 & 
 -0.671 & 0.997
\\ \hline
 G6 - G7 & 
 0.095 & 1.000
\\ \hline
 G6 - G2 & 
 0.314 & 1.000
\\ \hline
 G6 - G3 & 
 1.983 & 0.493
\\ \hline
 G6 - G4 & 
 -0.528 & 0.999
\\ \hline
 G6 - G8 & 
 -0.600 & 0.999
\\ \hline
 G7 - G2 & 
 0.186 & 1.000
\\ \hline
 G7 - G3 & 
 1.679 & 0.701
\\ \hline
 G7 - G4 & 
 -0.572 & 0.999
\\ \hline
 G7 - G8 & 
 -0.635 & 0.998
\\ \hline
 G2 - G3 & 
 1.694 & 0.692
\\ \hline
 G2 - G4 & 
 -0.843 & 0.991
\\ \hline
 G2 - G8 & 
 -0.932 & 0.983
\\ \hline
 G3 - G4 & 
 -2.476 & 0.206
\\ \hline
 G3 - G4 & 
 -2.656 & 0.136
\\ \hline
 G4 - G8 & 
 -0.042 & 1.000
\\ 
\hline
\end{tabular}
}
\label{tab:proof14a}
\end{table*}

\end{center}

\subsection{Evaluation for Statement in Figure \ref{fig:statement_data-AMT}(b)}
As a recap, Figure \ref{fig:statement_data-AMT}(b) evaluates the Likert responses to the following statement: ``Given the explanation, I am confident in the smart home's ability to accurately track Pat's activity". When conducting the one-way repeated ordinal regression with CLMM, we observe no significant effect of the 8 groups on the Likert responses ($\chi^2$(7) = 12.83, $p$=0.07). For sanity measures, we also report the pairwise comparison results from the Tukey test in Table \ref{tab:proof14b}, and observe no statistical differences between any pairs of groups.

\renewcommand{\arraystretch}{1}
\begin{center}
\begin{table*}[h]
\caption{Tukey Post-Hoc Analysis between Participant Groups for Figure\ref{fig:statement_data-AMT}(b)}
{\small
\begin{tabular}{|>{\centering\arraybackslash}p{1.5cm}|p{2.2cm}|p{2.2cm}|}
\hline
\multicolumn{1}{|c|}{\textbf{Pairwise Groups}} &\multicolumn{1}{c|}{\textbf{Z-ratio}} &\multicolumn{1}{c|}{\textbf{P-Value}} \\
 \hline
 G1 - G5 & 
 1.771 & 0.639
\\ \hline
 G1 - G6 & 
 2.090 & 0.422
\\ \hline
 G1 - G7 & 
 1.041 & 0.968
\\ \hline
 G1 - G2 & 
 0.000 & 1.000
\\ \hline
 G1 - G3 & 
 2.360 & 0.261
\\ \hline
 G1 - G4 & 
 0.040 & 1.000
\\ \hline
 G1 - G8 & 
 1.104 & 0.956
\\ \hline
 G5 - G6 & 
 0.255 & 1.000
\\ \hline
 G5 - G7 & 
-0.600 & 0.998
\\ \hline
 G5 - G2 & 
 -1.771 & 0.639
\\ \hline
 G5 - G3 & 
 0.480 & 0.999
\\ \hline
 G5 - G4 & 
 -1.672 & 0.705
\\ \hline
 G5 - G8 & 
 -0.728 & 0.996
\\ \hline
 G6 - G7 & 
 -0.853 & 0.989
\\ \hline
 G6 - G2 & 
-2.090 & 0.422
\\ \hline
 G6 - G3 & 
 0.230 & 1.000
\\ \hline
 G6 - G4 & 
 -1.974 & 0.499
\\ \hline
 G6 - G8 & 
 -1.018 & 0.972
\\ \hline
 G7 - G2 & 
 -1.041 & 0.968
\\ \hline
 G7 - G3 & 
 1.074 & 0.962
\\ \hline
 G7 - G4 & 
 -0.973 & 0.978
\\ \hline
 G7 - G8 & 
 -0.057 & 1.000
\\ \hline
 G2 - G3 & 
 2.360 & 0.261
\\ \hline
 G2 - G4 & 
 0.040 & 1.000
\\ \hline
 G2 - G8 & 
 1.104 & 0.9561
\\ \hline
 G3 - G4 & 
 -2.230 & 0.334
\\ \hline
 G3 - G4 & 
 -1.272 & 0.909
\\ \hline
 G4 - G8 & 
 1.023 & 0.971
\\ 
\hline
\end{tabular}
}
\label{tab:proof14b}
\end{table*}

\end{center}

\subsection{Evaluation for Statement in Figure \ref{fig:statement_data-AMT}(c)}
As a recap, Figure \ref{fig:statement_data-AMT}(c) evaluates the Likert responses to the following statement: ``The smart home provides adequate justification as to \textit{why} Pat is <activity>". When conducting the one-way repeated ordinal regression with CLMM, we observe no significant effect of the 8 groups on the Likert responses ($\chi^2$(7) = 4.08, $p$=0.76). For sanity measures, we also report the pairwise comparison results from the Tukey test in Table \ref{tab:proof14c}, and observe no statistical differences between any pairs of groups.
\renewcommand{\arraystretch}{1}
\begin{center}
\begin{table*}[h]
\caption{Tukey Post-Hoc Analysis between Participant Groups for Figure\ref{fig:statement_data-AMT}(c)}
{\small
\begin{tabular}{|>{\centering\arraybackslash}p{1.5cm}|p{2.2cm}|p{2.2cm}|}
\hline
\multicolumn{1}{|c|}{\textbf{Pairwise Groups}} &\multicolumn{1}{c|}{\textbf{Z-ratio}} &\multicolumn{1}{c|}{\textbf{P-Value}} \\
 \hline
 G1 - G5 & 
 1.059 & 0.965
\\ \hline
 G1 - G6 & 
 0.638 & 0.998
\\ \hline
 G1 - G7 & 
 0.873 & 0.988
\\ \hline
 G1 - G2 & 
 0.000 & 1.000
\\ \hline
 G1 - G3 & 
 0.856 & 0.989
\\ \hline
 G1 - G4 & 
 -0.357 & 1.000
\\ \hline
 G1 - G8 & 
 -0.164 & 1.000
\\ \hline
 G5 - G6 & 
 -0.435 & 0.999
\\ \hline
 G5 - G7 & 
-0.097 & 1.000
\\ \hline
 G5 - G2 & 
 -1.059 & 0.965
\\ \hline
 G5 - G3 & 
 -0.258 & 1.000
\\ \hline
 G5 - G4 & 
 -1.362 & 0.875
\\ \hline
 G5 - G8 & 
 -1.225 & 0.925
\\ \hline
 G6 - G7 & 
 0.299 & 1.000
\\ \hline
 G6 - G2 & 
-0.638 & 0.998
\\ \hline
 G6 - G3 & 
 0.193 & 1.000
\\ \hline
 G6 - G4 & 
 -1.964 & 0.9794
\\ \hline
 G6 - G8 & 
 -0.805 & 0.993
\\ \hline
 G7 - G2 & 
 -0.873 & 0.988
\\ \hline
 G7 - G3 & 
 -0.135 & 1.000
\\ \hline
 G7 - G4 & 
 -1.162 & 0.943
\\ \hline
 G7 - G8 & 
 -1.023 & 0.971
\\ \hline
 G2 - G3 & 
 0.856 & 0.989
\\ \hline
 G2 - G4 & 
 -0.357 & 1.000
\\ \hline
 G2 - G8 & 
 -0.164 & 1.000
\\ \hline
 G3 - G4 & 
 -1.183 & 0.937
\\ \hline
 G3 - G4 & 
 -1.030 & 0.970
\\ \hline
 G4 - G8 & 
 0.202 & 1.000
\\ 
\hline
\end{tabular}
}
\label{tab:proof14c}
\end{table*}

\end{center}

\section{Breakdown of Intraclass Correlation Coefficient - Section 6}
\label{sec:appendix-classifmapping}
Table \ref{tab:classifmapping-proof} provides a detailed breakdown of the intraclass correlation coefficient (ICC) for each activity. The ICC was calculated with a two-way random effects model, measuring an absolute agreement between the raters using the ratings from a single rater as the base measure, instead of the mean ratings of all raters. We utilized a two-way random effects model since it assumes that the selected raters are randomly selected from a population and rate all subjects (i.e. sensors for each activity).

\renewcommand{\arraystretch}{1}
\begin{center}
\begin{table*}[h]
\caption{Intraclass Correlation Coefficient Breakdown for Figure \ref{fig:exp_quality_proc}}
{\small
\begin{tabular}{|>{\centering\arraybackslash}p{2.2cm}|p{2.2cm}|}
\hline
\multicolumn{1}{|c|}{\textbf{Activity}} &\multicolumn{1}{c|}{\textbf{ICC}} \\
 \hline
Bathing & 
 0.86 \\
 \hline
Bed To Toilet & 
 0.87 \\
 \hline
Take Medicine & 
 1.0 \\
 \hline
Leave Home & 
 1.0 \\
 \hline
Work & 
 1.0 \\
 \hline
 Sleep & 
0.80 \\
 \hline
Cook & 
1.0 \\
 \hline
Relax & 
0.81 \\
 \hline
\end{tabular}
}
\label{tab:classifmapping-proof}
\end{table*}
\end{center}
\end{document}